\documentclass[10pt,journal,compsoc]{IEEEtran}

\usepackage{graphicx}
\usepackage{amsmath,amssymb}
\usepackage{color}
\usepackage{multirow}
\usepackage{booktabs}
\usepackage[T1]{fontenc}
\usepackage{subfigure}
\usepackage{float}
\usepackage{color}
\usepackage{setspace}
\usepackage{url}
\usepackage{makecell}

\ifCLASSOPTIONcompsoc
  \usepackage[nocompress]{cite}
\else
  \usepackage{cite}
\fi{\tiny {\tiny }}

\begin{document}

%% -----------------------------------------------------------------------------------------------
%% paper title
\title{Self-Regulated Learning for \\Egocentric Video Activity Anticipation}

\author{Zhaobo~Qi,~\IEEEmembership{Student~Member~IEEE,} Shuhui~Wang,~\IEEEmembership{Member~IEEE,}~Chi Su, \\~
Li Su,~Qingming~Huang,~\IEEEmembership{Fellow~IEEE,}
and~Qi~Tian,~\IEEEmembership{Fellow~IEEE}% <-this % stops a space
%\thanks{Corresponding author: Shuhui Wang.}
\IEEEcompsocitemizethanks{
	\IEEEcompsocthanksitem Corresponding author: Shuhui Wang and Qingming Huang.
	\IEEEcompsocthanksitem  Z. Qi, L. Su and Q. Huang are with the School of Computer Science and Technology, University  of  Chinese  Academy  of  Sciences, Beijing 101408, China, and with the Key Laboratory of Intelligent Information Processing, Institute of Computing Technology, Chinese Academy of Sciences, Beijing 100190, China. Q. Huang is also with Peng Cheng Laboratory, Shenzhen 518066, China. \protect\\
	E-mail:  zhaobo.qi@vipl.ict.ac.cn, \{suli, qmhuang\}@ucas.ac.cn.
	\IEEEcompsocthanksitem  S. Wang is with the Key Laboratory  of Intelligent Information Processing, Institute of Computing Technology, Chinese Academy of Sciences, Beijing 100190, China. \protect\\
	E-mail: wangshuhui@ict.ac.cn.
	\IEEEcompsocthanksitem C. Su is with Kingsoft Cloud, Beijing, 100085.\protect\\
	Email: suchi@kingsoft.com
	\IEEEcompsocthanksitem  Q. Tian is with Cloud BU, Huawei Technologies, Shenzhen 518129, China. \protect\\
	E-mail: tian.qi1@huawei.com. \protect}
} 

%% The paper headers
\markboth{Transactions on Pattern Analysis and Machine Intelligence}%
{Shell \MakeLowercase{\textit{et al.}}: Bare Demo of IEEEtran.cls for Computer Society Journals}

%% -----------------------------------------------------------------------------------------------
%% abstract or keywords.
\IEEEtitleabstractindextext{%
%!TEX root = ../article.tex

% Abstract
\begin{abstract}

Future activity anticipation is a challenging problem in egocentric vision. As a standard future activity anticipation paradigm, recursive sequence prediction suffers from the accumulation of errors. To address this problem, we propose a simple and effective Self-Regulated Learning framework, which aims to regulate the intermediate representation consecutively to produce representation that (a) emphasizes the novel information in the frame of the current time-stamp in contrast to previously observed content, and (b) reflects its correlation with previously observed frames. The former is achieved by minimizing a contrastive loss, and the latter can be achieved by a dynamic reweighing mechanism to attend to informative frames in the observed content with a similarity comparison between feature of the current frame and observed frames. The learned final video representation can be further enhanced by multi-task learning which performs joint feature learning on the target activity labels and the automatically detected action and object class tokens. SRL sharply outperforms existing state-of-the-art in most cases on two egocentric video datasets and two third-person video datasets. Its effectiveness is also verified by the experimental fact that the action and object concepts that support the activity semantics can be accurately identified.

\end{abstract}

}

\maketitle

\IEEEdisplaynontitleabstractindextext

\IEEEpeerreviewmaketitle

%% -----------------------------------------------------------------------------------------------
%% Sections
\IEEEraisesectionheading{\section{Introduction}\label{sec:introduction}}

Egocentric perception has received remarkable research attention in recent years. An increasing number of tasks~({\it e.g.}, egocentric video summarization, egocentric localization, egocentric object detection, egocentric action recognition and anticipation) and benchmark datasets~({\it e.g.},~EgoSum+gaze~\cite{xu2015gaze}, EGTEA Gaze+~\cite{li2018eye}, Charades-Ego~\cite{sigurdsson2018actor} and EPIC-Kitchens~\cite{damen2018scaling}) are proposed. The construction of large-scale egocentric video datasets, such as EPIC-Kitchens, further promotes the technical advance in this field. Among a diversified range of egocentric vision tasks, anticipating future activities, which aims to predict what will possibly happen in the future, has become an active research topic due to its wide application prospectives. For example, in human-robot interaction scenario, robots can work closely with humans if they are able to anticipate human actions in the next few minutes~\cite{koppula2015anticipating}, and in autonomous driving, an autonomous vehicle needs to anticipate if a pedestrian crosses the street and produce consequent system control command to guarantee driving safety~\cite{de2016online}.

\begin{figure}[t]
	\centering
	\subfigure[\textbf{Illustration of the problem.} Video delivers diversified content and rich context. Recursive sequence prediction model produces inaccurate intermediate representations.]{
		\includegraphics[width=0.95\linewidth]{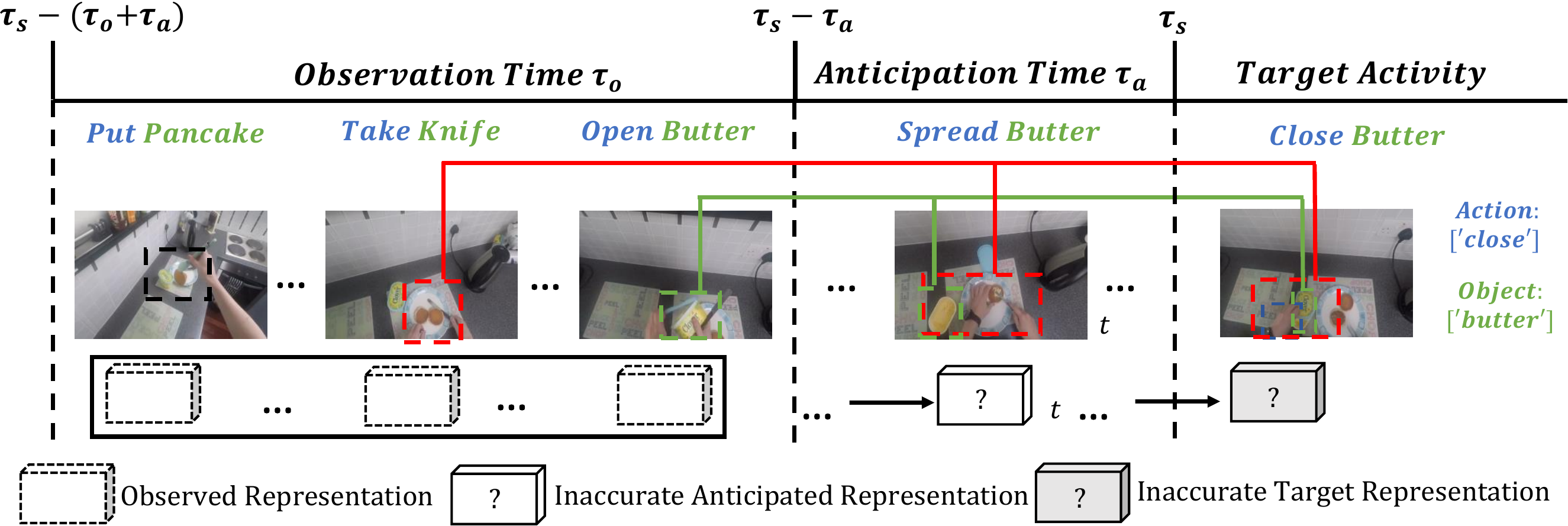}
		\label{fig:motivation-case.1}
		%\caption{fig1}
	}
	\quad
	\subfigure[\textbf{The workflow of SRL.} The predicted intermediate representation is rectified by Revision, the importance of previously observed key frame feature is reweighed by Reattend, and the semantic context information~(actions, objects) related to the target activity is used to enhance the target activity representation.]{
		\includegraphics[width=0.95\linewidth]{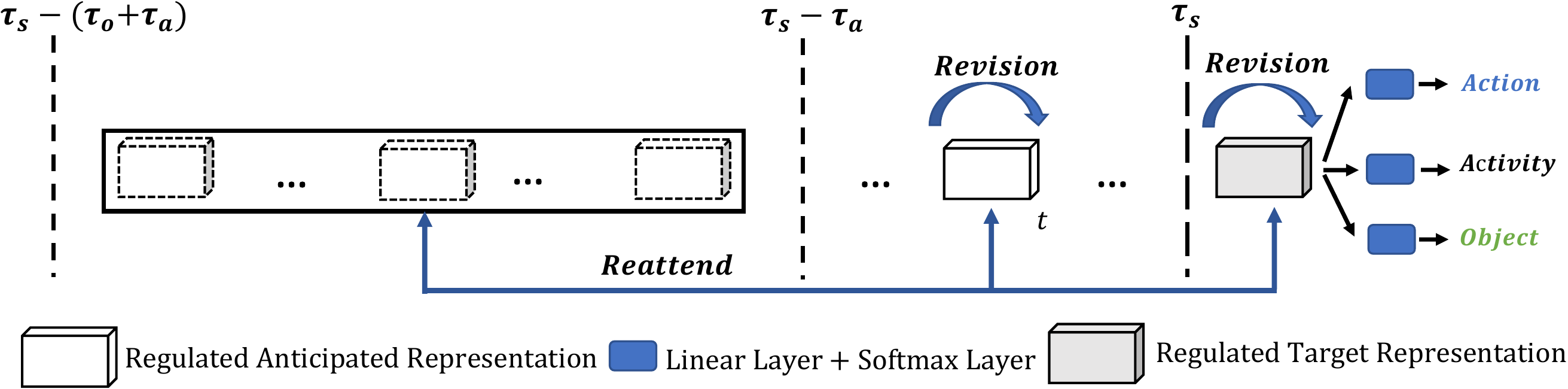}
		\label{fig:motivation-case.2}
		%\caption{fig2}
	}
	\quad
	\centering
	\caption{\textbf{Illustration of the problem and the major workflow of SRL}.}
	\vspace{-2ex}
	\label{fig:motivation-case}
\end{figure}

Due to recent progress in egocentric vision, existing models can predict what will happen within a time horizon of up to several minutes~\cite{abu2018will, farha2019uncertainty, ke2019time}. Nevertheless, future activity anticipation is still a challenging task. An example is shown in Figure \ref{fig:motivation-case.1}, as defined in~\cite{damen2018scaling}, the `anticipation time' $\tau_a$ is the time interval from the activity we need to anticipate, and the `observation time' $\tau_o$  is the observed length of the video in advance to the target activity. The goal of the future activity anticipation task is to predict the activity label of the video clip $[\tau_s, \tau_e]$ by observing a video clip $[\tau_s-(\tau_o + \tau_a), \tau_s-\tau_a]$, which precedes the target activity start time $\tau_s$ by $\tau_a$.

A standard future activity anticipation paradigm is the recursive sequence prediction~\cite{abu2018will, furnari2019would, farha2019uncertainty}, where the anticipation model sees all observed video clips and predicts what will happen at the next clip. The process is repeated until the desired prediction moment $\tau_s$ is reached, as shown in Figure~\ref{fig:motivation-case.1}. 
Based on this process, the key of obtaining accurate activity prediction at the end depends on how to extract and represent the informative visual cues in the video content during the anticipation stage.
In general, there are several crucial issues that need to be investigated.

First, activity videos contain drastic changes on both appearance and semantics from the beginning to the end. In Figure~\ref{fig:motivation-case.1}, the `make breakfast' video contains a series of activities, {\it i.e.}, `put pancake', `take knife', `open butter', `grab butter', `spread butter' and `close butter'. These activities are represented by key frames with very different appearance, {\it i.e.}, the actions and objects inside a frame may be different from one to another. On the other hand, different activities demonstrate consecutive but diversified contextual dependencies. In Figure~\ref{fig:motivation-case.1}, from object perspective, `butter' appears in frames of `open butter', `spread butter' and `close butter'; from event perspective, `spread butter' seems to have stronger contextual relation to `take knife'. 

At the beginning of recursive sequence prediction, we obtain an initial feature representation based on the observed video clip before $\tau_s-\tau_a$.
In model training, only the initial feature representations and the current frame in anticipating stage can be used, so the obtained consecutive intermediate representations may be far less accurate. 
If the intermediate representations are directly used in subsequent prediction, the accumulation of errors may result in inaccurate final target activity prediction. 
To address the above issues, we propose a simple and effective Self-Regulated Learning~(SRL) framework for future activity anticipation, which makes full use of the rich information only contained in the video to learn the anticipated representations in an unsupervised manner.

SRL aims to regulate the intermediate representation consecutively to produce representation that (a) emphasizes the novel information at the current time-stamp in contrast to previously observed content; and (b) reflects its correlation with previously observed frames. 
See Figure~\ref{fig:motivation-case.2},
the former requires a revision operation on the intermediate representation generated by the pre-trained sequential prediction model. At each anticipation time-step, a contrastive loss is utilized to rectify the predicted intermediate feature representation by treating the intermediate representation as positive sample and a batch of semantically uncorrelated frames as negatives sampled without difficulty.
The latter demands a dynamic reweighing mechanism to attend to informative frames in the observed content, which resorts to a similarity comparison between feature of the current frame and observed frames. The highly similar frames may dominate the reweighed observed features, which, in combination with the revised representation, is fed into another sequential prediction unit to produce a sequence of more complete and accurate feature representations. 

Compared to recent iterative prediction methods~\cite{abu2018will, furnari2019would, farha2019uncertainty}, SRL can learn a representation that is less error-prone and avoids performance degradation due to accumulation of anticipation error. Compared to other works ~\cite{mahmud2017joint, ke2019time} that directly utilize observed representations to anticipate long-term future activity without intermediate anticipation, the intermediate representations learned by SRL take full advantage of the rich context in videos. The learned final video representation can be further improved by considering the rich semantic context information by exploring the mid-level semantic tokens, {\it e.g.}, the activity `close butter' contains action `close' and object `butter'. These tokens can be easily identified from the activity description labels by automatically detecting the nouns and verbs, respectively. 
Finally, we employ a multi-task learning framework to perform joint feature learning on the targe activity labels and the detected action and object class tokens.

We carry out extensive experiments on two egocentric video datasets~(EPIC-KITCHEN and EGTEA Gaze+) to verify the effectiveness of our proposed model. 
The proposed SRL is also evaluated on third-person video datasets~(50 Salads~\cite{stein2013combining} and Breakfast~\cite{kuehne2014language}) to prove the generality of our model in predicting future activities. 
Experiments show that our method achieves promising performances, which sharply outperforms existing state-of-the-art in most cases on the four benchmark datasets. Source codes are available at~\url{https://github.com/qzhb/SRL}.

The contribution of our work are three folds:
\begin{itemize}
	\item We propose SRL, a self-regulated learning framework for egocentric activity anticipation. It recursively produces more accurate intermediate representations at any anticipation time-step by iteratively rectifying the current visual representation and reattending to the most relevent observed video frames. 
	\item By exploring mid-level semantic tokens from actions and objects with multi-task learning framework, a more semantically enhanced and self-regulated target representation is obtained for target activity anticipation.
	\item SRL achieves competitive performance on both egocentric and third-person video datasets. The effectiveness of SRL is also verified by the experimental fact that it can accurately identify action and object concepts that explain the activity semantics.
\end{itemize}

%\clearpage
\section{Related Work}
\label{sec:related-work}

We briefly review recent advances in egocentric video analysis in Section~\ref{subsec:related-egocentric}. We further discuss related approaches in third-person video analysis in Section~\ref{subsec:related-third-person}. %Besides, we also discuss the related work about contrastive predictive coding in Section~\ref{subsec:related-cpc}.

\subsection{Egocentric Video Analysis}
\label{subsec:related-egocentric}

\subsubsection{Egocentric Video Recognition}
Egocentric video recognition has been undergoing speedy development in recent years, evidenced by the emergence of many new large-scale egocentric video datasets ~\cite{xu2015gaze},~\cite{li2018eye}, \cite{sigurdsson2018actor}, \cite{damen2018scaling}, and a huge body of research work found in literature ~\cite{fathi2011understanding,fathi2012learning,li2018eye,ma2016going,ryoo2015pooled,singh2016first,spriggs2009temporal,sudhakaran2019lsta,sudhakaran2018attention,fathi2011learning,poleg2016compact}.

As an early endeavor, Spriggs {\it et al.} \cite{spriggs2009temporal} utilize a wearable camera and Inertial Measurement Units (IMUs) to explore first-person sensing and perform temporal segmentation and classification of human activity. Considering the strong contextual relation shown in first-person video, Fathi {\it et al.}~\cite{fathi2011understanding} exploit the consistency of appearance representation of actions, hands, and objects, and propose a hierarchical activity modeling framework. 
Later, it was found that gaze location is a very important clue for egocentric video recognition, so the gaze location is first used for identifying salient visual information in~\cite{fathi2011learning}. 
A probabilistic generative model is proposed to learn the spatio-temporal relationship between gaze point, scene objects, and activity label in first-person video for daily activity recognition in~\cite{fathi2012learning}.

With the development of deep learning, CNN is employed in~\cite{ryoo2015pooled, poleg2016compact} for video feature representation in egocentric video recognition. Ma {\it et al.}~\cite{ma2016going} design a CNN-based two-stream network to integrate appearance and motion for egocentric activity recognition. Li {\it et al.}~\cite{li2018eye} consider the gaze as a probabilistic variable and use a deep network to model its distribution for joint egocentric video recognition and gaze prediction. 
Sudhakaran {\it et al.}~\cite{sudhakaran2019lsta} propose Long Short-Term Attention (LSTA), a new recurrent neural unit to pay attention to features from relevant spatial parts for egocentric activity recognition. 

The rapid development of activity recognition deepens our understanding of egocentric video, and also benefits research on other related topics, such as egocentric video summarization, egocentric object detection and egocentric video anticipation. 

%We will next review the developments of egocentric video anticipation.

\subsubsection{Egocentric Video Anticipation} 

Activity anticipation for egocentric vision has been extensively studied in~\cite{damen2018scaling,ke2019time,furnari2019would,qi2017predicting,furnari2018leveraging,miech2019leveraging,wu2017anticipating,bokhari2016long,fan2018forecasting,furnari2017next,soo2016egocentric,rhinehart2017first,ryoo2015robot,singh2016krishnacam,soran2015generating,zhang2017deep}.
Some works focus on predicting the next activity. 
Qi {\it et al.}~\cite{qi2017predicting} propose a spatial-temporal And-Or graph (AOG) to represent events, and an early parsing method using temporal grammar is established to anticipate the next activity.
Others focus on predicting what will happen after a long time interval. 
Ke {\it et al.}~\cite{ke2019time} propose to explicitly condition the anticipation on time, which is shown to be efficient and effective for long-term activity anticipation. 
Furnari {\it et al.}~\cite{furnari2017next} propose to explore the dynamics of the scene and introduce a model to analyze fixed-length trajectory segments to forecast the next active objects.
 
Recently, a new egocentric activity anticipation challenge is proposed in~\cite{damen2018scaling}.
On this challenging dataset, Furnari {\it et al.}~\cite{furnari2018leveraging} propose new loss functions for activity forecasting, which incorporate the uncertainty of the prediction of future activities. 
A learning architecture, {\it i.e.}, RU-LSTM~\cite{furnari2019would}, is proposed, which processes RGB, optical flow and object-based features using two LSTMs and a modality attention mechanism to anticipate future activities. 

Most of the above activity anticipation techniques have not taken full advantage of the rich context in the video content in their models. Besides, the semantic context information existing among the target activity is left to be uninvestigated. 
In comparison, we develop a self-regulated learning process which uses the contextual relation along the temporal direction in video sequences on both feature and semantic level to adaptively refine the video features for activity anticipation.
Our model can be applied to different activity anticipation settings, like long time anticipation as in~\cite{abu2018will} and the challenge in~\cite{damen2018scaling} on large-scale datasets.

\subsection{Third-person Video Analysis}% and Prediction}
\label{subsec:related-third-person}

\subsubsection{Third-person Video Recognition} 
Numerous works based on deep CNN have been proposed for video action recognition. 
For typical 2D-CNN-based methods~\cite{karpathy2014large}, 2D convolution is simply applied on single video frame and the frame features are fused. To model temporal information, two-stream-based methods~\cite{feichtenhofer2016convolutional, feichtenhofer2017spatiotemporal, wang2016temporal} are introduced to model appearance and dynamics separately with two networks and they are fused in the middle stage or at the end. 
In another branch, many 3D-CNN-based methods (C3D~\cite{tran2015learning}, I3D~\cite{carreira2017quo}, ECO~\cite{zolfaghari2018eco} and SlowFast~\cite{feichtenhofer2019slowfast}) have been proposed to learn spatio-temporal features from RGB frames directly. Besides, the concept representations are utilized to perform video recognition~\cite{qzhb2020ckmn, qzhb2020tdc}, which is believed to provide better interpretability.

Action recognition models, which can efficiently extract video feature representations, provide backbone models for the development of egocentric video understanding. For example, the TSN model~\cite{wang2016temporal} is utilized to extract observed video clip representations for egocentric activity anticipation in RU-LSTM~~\cite{furnari2019would}. In our model, the TSN and I3D model are used as video feature extractor.

\subsubsection{Third-person Video Anticipation and Prediction} 
%In addition to the above content, other advances in video analysis are also closely related to activity anticipation. We briefly review the most related.

Third-person video anticipation predicts future activity categories from the third-person perspective~\cite{lan2014hierarchical, abu2018will, felsen2017will, jain2016recurrent, mahmud2017joint, kitani2012activity, sadegh2017encouraging}. 
Mahmud {\it et al.}~\cite{mahmud2017joint} propose a hybrid Siamese network for jointly predicting the label and the starting time of future unobserved activity.
Farha {\it et al.}~\cite{abu2018will} propose an RNN-based model and a CNN-based model to obtain accurate predictions, which scale well on different datasets and videos with varying lengths and huge variations in the possible future actions. 

Third-person action prediction, also referred to as early stage action recognition, aims at predicting the label of an action as early as possible from partial observations~\cite{de2016online}. Much effort has been devoted to action prediction~\cite{ryoo2011human, de2016online, sadegh2017encouraging, gao2017red, kong2017deep, shi2018action, liu2019skeleton}. 
As one of the earliest works on this problem, Ryoo {\it et al.}~\cite{ryoo2011human} use dynamic visual bag-of-words to model changes in feature distribution over time. 
Shi {\it et al.}~\cite{shi2018action} first try to obtain future visual feature through regression, and then an action recognition model is used for action prediction.  
Kong {\it et al.}~\cite{kong2017deep} propose a deep sequential context network to reconstruct missing information of the partial observed video for action prediction. Compared to first-person videos, the semantic relation between temporally adjacent video segments appears to be more diversified.

%\clearpage
\begin{figure*}[ht]
	\begin{center}
		\includegraphics[scale=0.52]{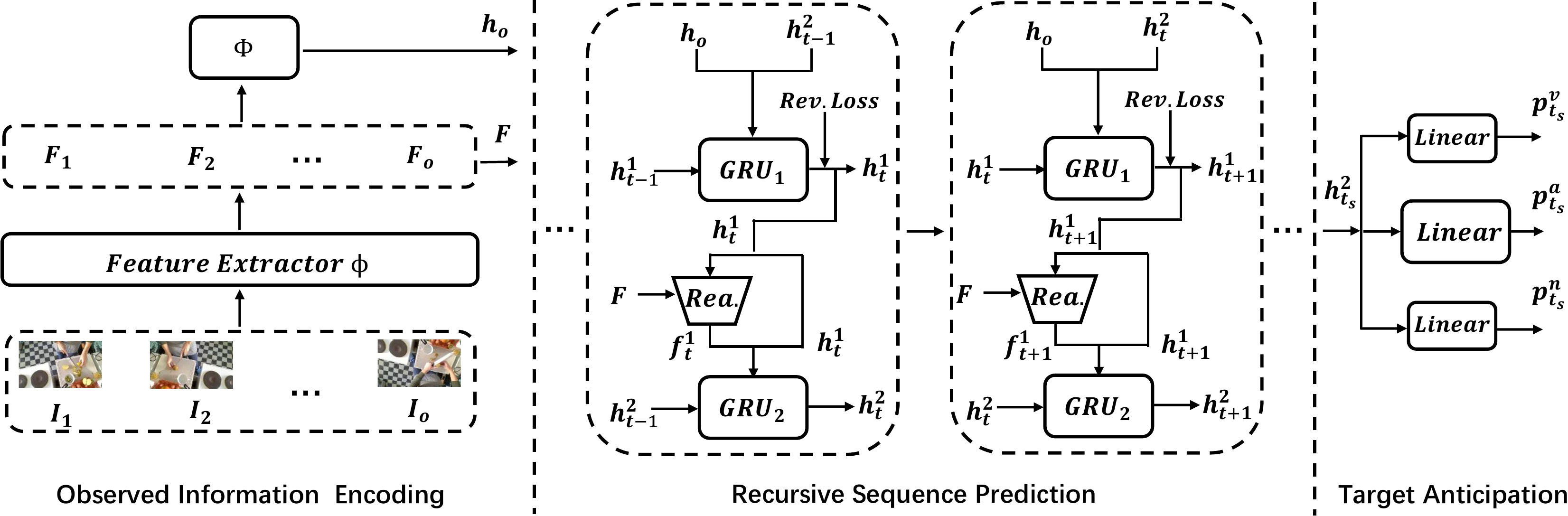}
	\end{center}
	\caption{The proposed \textit{\textbf{SRL}} framework. SRL consists of three main steps for future activity anticipation. In the observed information encoding step, given the observed video clip $I$, a feature extractor $\phi$ and an aggregation function $\Phi$ are employed to obtain feature representations $\boldsymbol{F}$ at each observed time-step and hidden representation $h_o$ at the last observed time-step. At the recursive sequence prediction step, a $GRU$ layer is utilized to obtain the initial feature representation $h^1_t$, then the $Rev.$ loss is employed to rectify it. After that, the revised representation and the observed representation $\boldsymbol{F}$ are fed into the $Rea.$ module to obtain representation $f^1_t$ that relates to current video content. At last, $f^1_t$ and $h^1_t$ are fused by another $GRU$ layer to get the final representation at current time-step. In the target anticipation step, a multi-task learning framework is utilized to enhance the final representation by exploiting the semantic context information~(actions $p^v_{t_s}$ and objects $p^n_{t_s}$) related to the target activity, and the predicted activity probability distribution $p^a_{t_s}$ is obtained.}
	\label{fig:framework}
	%\vspace{-2ex}
\end{figure*}

\section{Approach}
\label{sec:methods}

Our model SRL is developed upon the general recursive prediction framework, see Figure~\ref{fig:framework}. We first describe the general framework in Section~\ref{subsec:anticipation-framework}. Then we show how to perform self-regulated learning in details and discuss how the challenges described in Section~\ref{sec:introduction} are addressed in Section~\ref{subsec:methods-revision}, Section~\ref{subsec:methods-reattend} and Section~\ref{subsec:methods-context}. Finally, we give our overall learning objective function in Section~\ref{subsec:methods-objective-function}.

\subsection{Future Activity Anticipation Framework}
\label{subsec:anticipation-framework}

The target of future activity anticipation is to predict the activity label of a video clip starting at time $\tau_s$ by observing a video clip starting at $\tau_s-(\tau_o + \tau_a)$ and ending at $\tau_s-\tau_a$, which precedes the target activity start time by $\tau_a$. In other words, we need to predict what will happen after $\tau_a$ by observing a video clip of length $\tau_o$. For simplicity, similar as~\cite{furnari2019would}, we extract video frames every $\delta$ seconds on both the observed part and to-be-anticipated part. Therefore, we assume that the observed video clip contains $o$ frames, represented as $I=\left\{{I_1, I_2, ..., I_{o}}\right\}$, and the anticipation process contains $a$ frames. We use $t$ to index the current frame to be anticipated and $t_s$ to index the target frame to be anticipated. 

As shown in Figure~\ref{fig:framework}, SRL contains three main steps, {\it i.e.}, the observed information encoding step, the recursive sequence prediction step and the target activity anticipation step. Now we introduce each step in detail.

\subsubsection{Observed Information Encoding}

For activity anticipation, all the available information we can get for prediction is obtained from the observed video clip. Therefore, how to effectively encode the observed video clip is the foundation for subsequent recursive sequence prediction process.
As shown in Figure~\ref{fig:framework}, given the observed video clip $I=\left\{{I_1, I_2, ..., I_{o}}\right\}$, first a feature extractor $\phi$ is utilized to obtain the feature representation $F_j\in{\mathbb{R}^d}$ of the observed video frame at each time-step $j$. We can use many base models as $\phi$, ${\it e.g.}$, the TSN model~\cite{wang2016temporal} and the I3D model~\cite{carreira2017quo}.
Then we use an aggregation function $\Phi$~(such as RNN model) to encode the observed video representations, and obtain the hidden representation $h_o\in{\mathbb{R}^d}$ at the last observed time-step. The process is shown as 
\begin{equation}
F_j = \phi(I_j); \ \ h_o = \Phi(\left\{F_1, F_2, ..., F_{o}\right\})
\end{equation}
%\phi \varphi \Phi

The obtained representation $\boldsymbol{F} = \left\{F_1, F_2, ..., F_{o}\right\}$ and $h_o$ will be utilized in the following recursive sequence prediction process. The choice of $\phi$ and $\Phi$ will be given in Section~\ref{sec:experiment-setting} and~\ref{subsubsec:aggregation_functions}.

\subsubsection{Recursive Sequence Prediction}

For recursive sequence prediction, given the observed video representation $\boldsymbol{F}$ and the hidden representation $h_o$ at the last observed time-step, we predict what will happen at the next anticipation time-step repeatedly until the target anticipation time-step is reached. 

As mentioned in Section~\ref{sec:introduction}, due to error accumulation of recursive sequence prediction, the anticipated intermediate feature representation may be inaccurate. How to utilize the diversified content and rich context contained in the video to regulate the predicted representation is the core issue at the anticipation process for obtaining an accurate and complete intermediate representation. 

%Now we will show how to take these two factors into full account in the iterative prediction process.

Specifically, in Figure~\ref{fig:framework}, at each anticipation time-step, given the feature representation $h_o$ at the last observed time-step and the predicted feature representation $h^2_{t-1}$ at the last anticipation time-step, a $GRU$ layer~($GRU_1$) is first employed to obtain the initial prediction feature representation $h^1_{t}$ at this anticipation time-step as
\begin{equation}
h^1_t = GRU_1([h_o, h^2_{t-1}], h^1_{t-1})
\end{equation}
Applying the initial feature representation directly for subsequent predictions will lead to error accumulation and inaccurate final anticipation results. To get a more accurate feature representation, we utilize the contrastive loss function to rectify the predicted representation. The detail analysis will be shown in Section~\ref{subsec:methods-revision}.

%This initial feature representation is not accurate for anticipation task. 

After the representation revision process, a representation on the video content that is expected to be more accurate at this anticipation time-step is obtained. The next step is to obtain useful information from the observed video clip that is related to the current video content. As shown in Figure~\ref{fig:framework}, the representation $h^1_{t}$ will be used to dynamically attend to the observed video representation and acquire useful information $f^1_{t}$ related to $h^1_{t}$. Then $f^1_{t}$ and $h^1_{t}$ are fused to get the final intermediate representation $h^2_{t}$, which will be used to anticipate what will happen. We will give detail analysis in Section~\ref{subsec:methods-reattend}. 

Finally, we perform the above procedures iteratively until the target anticipation time-step is reached.

\subsubsection{Target Activity Anticipation}

After the recursive sequence prediction, we can obtain the representation $h^2_{t_s}$ at the final anticipation time-step $t_s$. Next, we describe in detail how to model semantic context information related to the target activity for obtaining the final accurate activity prediction results in Section~\ref{subsec:methods-context}.

The ultimate goal of our model is to get the activity categories at the target anticipation time-step. The probability distribution of the target activity $p^a_{t_s}\in{\mathbb{R}^{N_a}}$ at the final time-step can be calculated by a linear layer with softmax activation function,
\begin{equation}
p^a_{t_s}  = softmax(\boldsymbol{W}_{a}\hat{h}^2_{t_s}+b_a)
\end{equation}
where $\boldsymbol{W}_{a}\in{\mathbb{R}^{d\times{N_a}}}$ is the learnable parameters, and $N_a$ is the number of activity categories. $\hat{h}^2_{t_s}$ is the concatenation of $h^2_{t_s}$ and $h^1_{t_s}$, which increases the representation capability. In our work, we optimize the cross entropy loss $L_{a}$ to train the activity anticipation model.

\subsubsection{Multiple Future Activities Prediction}
Once the SRL has been trained, we predict what will happen at multiple future moments following a recursive sequence anticipation style. Given an observed video clip, our model acquires the feature representation of the clip. At each anticipation time-step $t$, we can get the final feature representation $h_t^2$ through the Recursive Sequence Prediction module. $h_t^2$ will be used to obtain the activity label that is taking place at this time-step through the Target Activity Anticipation module. On the other hand, $h_t^2$ is again forwarded through the Recursive Sequence Prediction module and Target Activity Anticipation module to produce the next prediction. The anticipation results at multiple future time-steps are obtained by repeatedly forwarding the previously generated prediction through the Recursive Sequence Prediction module and Target Activity Anticipation module until the desired final moment is reached.

\subsection{Representation Revision with Contrastive Loss}
\label{subsec:methods-revision}

% One intuitive idea is to correct the anticipated representation element by element by minimizing loss functions such as mean-square error. Besides, element level corrections also create computational overhead. 

% 1. X怎么采样，在recursive过程中，这个采样集合怎么变化才能保证算法效果好。
% 2. 通过对比损失最小化，突出了什么？是不是与当前sub-event语义紧密相关的那些信息？
% 3. 这个方法实际上是一种sub-event层面的semantic boosting。
% 4. 其他人是否做过类似的事情，他们针对的问题是什么，目标是什么，和本研究的异同在哪里

At each anticipation time-step, the obtained initial representation $h^1_{t}$ needs to be revised to adapt to the anticipation at time-step $t$. For a long video, due to the semantic coherence between adjacent video frames, the overall high-level semantic information of a video content should be consistent along the time. 
For example, the video `make breakfast' in Figure~\ref{fig:motivation-case} contains multiple activities. These series of activities are closely related but distinctive. They can be accurately identified by feature representations containing higher-level semantics. Ideally, for an activity anticipation model, the ability to capture event semantics in anticipation stage is necessary for accurate prediction. 

Unfortunately, there is no event label on the video sub-sequence to be anticipated. 
As an unsupervised learning paradigm, contrastive loss~\cite{hadsell2006dimensionality} has been widely used in audio and image recognition tasks. It is able to optimize the similarity of sample pairs in the feature space~\cite{oord2018representation, henaff2020data} for unsupervised representation learning on high-dimension data. 

We apply contrastive loss on our video anticipation task to regulate the predicted feature representation only based on the video content. 
By using this loss, the representational ability of anticipated features can be enhanced by enforcing the difference between features of different sub-events.
A recently proposed contrastive loss function InfoNCE~\cite{oord2018representation} is used in our model. The basic idea is to form a binary classification task that can correctly distinguish the target among a set of samples.

Specially, at each anticipation time-step, given the feature representation set $ X = \left\{ x_t^0, ..., x_t^{N-1} \right\} $ with $N$ samples and the initial prediction feature representation $h^1_t$, the contrastive loss function can be expressed as
\begin{equation}
L^t_{rev} = -\mathop{E}\limits_{X}\Big [\log \frac{\exp(h^1_t * x^0_t)}{\sum_{x^j_t\in{X}}{\exp(h^1_t * x^{j}_t)}} \Big ]
\end{equation}
where $*$ represents dot product. By minimizing this loss function, we can obtain a revised intermediate feature representation $h_t^1$.

The set $X$ contains one positive sample that is the feature representation $x^0_t$ at this anticipation time-step, and $N-1$ randomly sampled negative samples $\left\{x^1_t, ..., x^{N-1}_t\right\}$. In our method, $x^0_t$ is obtained by inputting the anticipation frame into the feature extractor $\phi$. For the negative samples, to ensure the effectiveness of the representation revision operation, it is expected that the negative samples contain samples with similar (but different) semantic information and with different semantic information to the target sample, where the former can be treated as `hard negatives'. We split each video into multiple clips according to the activity labels. Each clip is used as a training instance. We sample negative samples randomly from video clips with different activity labels as the selected negative training set. These negative samples for calculating the contrastive loss may come from videos that have the same or different video ids as the positive sample.  This setting can better guarantee the diversity and the similarity of the negative samples compared to positive sample. We will give detail analysis on the number of samples $N$ and the sampling methods in Section~\ref{subsubsec:baseline_rev}.

%For example, the positive sample and negative samples are recorded by the same actors in the same kitchen scene or different actors and kitchen scenes.

% For the $N-1$ negative samples, to ensure its diversity, we use the features of the randomly sampled $N-1$ frames from other videos to ensure that they have different sub-event label with the to-be-anticipated frame.

%,and then put them into $\phi$ to obtain the feature representations.

\subsection{Dynamically Reattending and Fusion}
\label{subsec:methods-reattend}

%Therefore, we had better to efficiently utilize this useful information, which can be used to enhance the representation capability of the predicted feature and help us make accurate prediction in the anticipation process.

For activity videos with varied time duration, the contents of key frames have noticeable correlation among one another. For example, in Figure \ref{fig:motivation-case.1}, the activity `spread butter' at anticipation time-step $t$ is closely related to the objects `pancake', `knife', `butter' and the action `open butter' in frames that appear in previously observed video clip.
Correspondingly, given a frame to be anticipated, the importance of frames in the observed video clip should be reweighed to enforce those truly related frames. Also, at different anticipation time-steps, the importance of observed frames should be adaptively adjusted due to the content change.

%To further enhance the adaptation ability of the recursive sequence prediction process, 

We design a module performing dynamic reattending on the frames in the observed video clip for activity anticipation. 
Given $h^1_{t}\in\mathbb{R}^d$ at anticipation time-step $t$ and the representation $\boldsymbol{F}\in\mathbb{R}^{d\times{o}}$ of the observed video clip, we define a similarity vector $s_t=\left\{{s^1_t, ...,s^j_t, ..., s^o_t}\right\}\in{\mathbb{R}^o}$, which represents the correlation between the feature at the current time-step and those at each observed time-steps. For example, $s^j_t$ indicates how much useful information we can get from video content at observed time-step $j$. $s^j_t$ can be calculated as follows,
\begin{equation}
s^j_t = \frac{F_j*h^1_t}{||F_j||||h^1_t||}
\end{equation}
where $*$ represents dot production.
Then we use $s^j_t$ to reattend to the useful observed information to get $f^1_t$ by
\begin{equation}
f^1_t = \sum^o_{j=1}s^j_tF_j
\end{equation}

%which can be used to predict what happens at current time-step. The process of this layer is,

Recall that $h^1_t$ expresses the video content of the current time-step, and the reattended representation $f^1_t$ contains more useful relevant information in the past, these two representations can complement each other effectively. To make full use of them, we use another $GRU$~($GRU_2$) layer to obtain a more complete feature representation $h^2_t\in{\mathbb{R}^d}$ as
\begin{equation}
h^2_t = GRU_2([h^1_t, f^1_t], h^2_{t-1})
\end{equation}

\subsection{Semantic Context Exploration}
\label{subsec:methods-context}
%Therefore, how to efficiently utilize this representation to get the precise activity is the next core problem.

After recursive sequence prediction process with representation revision and reattending, we have obtained $h^2_t$ that is expected to be more accurate and more comprehensive for target activity anticipation. Besides, for a target activity, there is some useful semantic context information. As shown in Figure~\ref{fig:motivation-case.1}, the activity `close butter' can be described by mid-level semantics such as action\footnote{The action is described by verb in the activity label.} `close' and object\footnote{The object is described by the nouns in the activity label.} `butter' that tells the subject and object of the target activity. We refer to these activity-related actions and objects as semantic context. Thus, it is helpful to employ these semantic contexts to make better activity prediction. 

Given the main task, {\it i.e.},  the target activity anticipation by minimizing loss $L_{a}$, we construct two auxiliary tasks for action categorization and object categorization by minimizing their respective cross-entropy loss $L_v$ and $L_n$. The three tasks are performed under the
multi-task learning framework, which has been
shown to improve the model generality and performance on the main task~\cite{caruana1997multitask,sener2018multi,kapidis2019multitask}. 

%the target anticipation time-step representation 

Specially, given $h^2_{t_s}\in{\mathbb{R}^d}$, we use two separate linear layers with softmax to predict the probability distribution of the related actions and objects,
\begin{equation}
\begin{split}
&p^v_{t_s}  = softmax(\boldsymbol{W}_{v}\hat{h}^2_{t_s}+b_v)\\
&p^n_{t_s}  = softmax(\boldsymbol{W}_{n}\hat{h}^2_{t_s}+b_n)\\
\end{split}
\end{equation}
where $\boldsymbol{W}_{v}\in{\mathbb{R}^{N_v \times N_d}}$ and $\boldsymbol{W}_{n}\in{\mathbb{R}^{N_n \times N_d }}$ are learnable parameters and $N_v$ and $N_n$ are the numbers of categories of actions and objects, respectively. $\hat{h}^2_{t_s}$ is the concatenation of $h^2_{t_s}$ and $h^1_{t_s}$, which increases the representation capability. Together with minimizing $L_{a}$, the two cross-entropy loss functions $L_{v}$ and $L_{n}$ are minimized to learn the two linear layers.

\subsection{Training Objective Function}
\label{subsec:methods-objective-function}

Given the parameters $\theta$ of our model, the overall training objective function can be expressed as,
\begin{equation}
L(\theta) = L_{a} + \alpha \cdot (L_{n} + L_{v}) + \beta \cdot \sum^{a}_{t=1}L^t_{rev}
\label{objective-function}
\end{equation}
where $\cdot$ is scale multiplication. $0 \le \alpha \le 1$ and $0 \le \beta \le 1$ are the weights of the corresponding loss function.  $L_{a}$ is an entropy loss for final activity classification, and $L_{n}$ and $L_{v}$ are the loss functions for action and object classification of the multi-task learning task. In addition, the third term in Equation~\ref{objective-function} is the sum of the contrastive loss at all anticipation time. Our model can be trained end-to-end.

%, and by optimizing the above loss function, accurate activity prediction results can be obtained.

%\clearpage
\section{Experiments}
\label{sec:experiments}

We evaluate SRL on both egocentric and third-person video datasets to verify the general applicability for future activity anticipation. %We also perform experiments on  of our proposed method. %The detail settings and results are described as follows. 

\subsection{Datasets and Metric}
\subsubsection{Datasets}
\textbf{EPIC-Kitchens Dataset~\cite{damen2018scaling}} is a large scale cooking video dataset from a first person view captured by 32 subjects in 32 different kitchens. Each video is composed of multiple activity segments, annotated with 125 action and 352 object classes. There are 272 video sequences with 28561 activity segments for training/validation and 160 video sequences with 11003 activity segments for testing. Since the annotations of the test videos are not available, following~\cite{furnari2019would}, we split the training set into training and validation sets by randomly choosing 232 videos for training and 40 videos for validation. We consider all unique (action, object) class pairs in the public training set, and obtain 2513 unique activity classes. We also report results on the test set with seen~(\textbf{S1}) and unseen~(\textbf{S2}) kitchens. \textbf{S1} indicates the test set includes scenes appearing in the training set, and \textbf{S2} means the test set includes scenes not appearing in the training set. 
% for training $L_v$.

% {\bf Note: How many objects for all the four datasets?}
%\textbf{Charades-Ego Dataset \cite{sigurdsson2018charades}.} This dataset is a large-scale dataset of paired third and first person videos. There are 157 activities and 68,536 activity instances in the dataset. Totally, there are 34.4 hours of third person video and another 34.4 hours of corresponding first person videos. The videos are splitted into 80/20 training set (6167 videos) and test set (1693 videos). In our experiments, we only use the first person videos.

\textbf{EGTEA Gaze+ Dataset~\cite{li2018eye}} contains 28 hours of first person cooking activity videos from 86 unique sessions of 32 subjects performing 7 meal preparation tasks. Each video contains audios, gaze tracking, human annotations of activities and hand masks. This dataset includes 10325 instances of activities, 19 action classes, 51 object classes and 106 unique activity classes. Three different train/test splits are provided by the authors, and we report the average performance of our model across all three splits. 

\textbf{50 Salads Dataset~\cite{stein2013combining}} contains 50 videos of salads preparation activities which are performed by 25 actors. 
%Each actor prepares two mixed salads. 
The dataset is composed of 17 fine-grained activity classes. As no action and object classes are provided, we decouple 7 unique action classes and 14 object classes from all the activity categories. Following~\cite{stein2013combining}, we utilize a five-fold cross-validation for evaluation.

\textbf{Breakfast Dataset~\cite{kuehne2014language}} is composed of 1712 videos of people preparing breakfast meals. It contains 48 fine-grained activity categories. Similar to 50 Salads dataset, we also decouple 15 unique action classes and 36 object classes from all the activity categories. The videos are recorded in 18 different kitchens containing 52 different actors from third person view. The dataset is split into four different train/test splits: S1, S2, S3 and S4. We use these four splits for evaluation. 

\subsubsection{Metric}
For EPIC-Kitchens, following~\cite{damen2018scaling}, we use the Top-5 accuracy as a class-agnostic measure and Mean Top-5 Recall as a class-aware metric. Specifically, Mean Top-5 Recall is averaged over the provided list of many-shot actions, objects and activities. For the EPIC-Kitchens test set, we use the official evaluation metrics, {\it i.e.}, Top-1 accuracy, Top-5 accuracy, Average Class Precision and Average Class Recall. 
For EGTEA Gaze+, Top-5 accuracy is used as the evaluation criterion. For 50 Salads and Breakfast, we use mean accuracy over classes for performance comparison.

\subsection{Implementation Details}

\subsubsection{Experiment Settings}
\label{sec:experiment-setting}

For EPIC-Kitchens and EGTEA Gaze+, all video clips are processed every 0.25s. The input of our model is a fixed-length video clip~({\it i.e.}, 1.5s in our experiments), and the goal is to anticipate what will happen at multiple time-steps~({\it i.e.}, 0.25s, 0.5s, 0.75s, 1s, 1.25s, 1.5s, 1.75s and 2s). In other words, the observed time-step $o$ is 6 and the anticipation time-step $a$ is 8. 
%In our model, all video clips are processed every 0.25s. For EPIC-Kitchens and EGTEA Gaze+, the input of our model is a fixed-length video clip~(2s in our experiments), and the goal is to anticipate what will happen at next multiple time-steps~(0.25s, 0.5s, 0.75s, 1s, 1.25s, 1.5s, 1.75s and 2s in our experiments). In other words, the observed time-step $o$ is 6 and the anticipation time-step $a$ is 8.
For 50 Salads and Breakfast, we follow the dense anticipation protocol in \cite{abu2018will} for the convenience of comparison with other methods. In this setting, the input is a particular percentage ({\it i.e.}, 20\% and 30\%) of each video, and the goal is to anticipate the activity labels of the following sub-sequence with a percentage ({\it i.e.}, 10\%, 20\%, 30\% and 50\%) of the video. 

For aggregation function $\Phi$, we use a simple GRU layer. Other aggregation function like average pooling can also be utilized in our model. We will give detail analysis of different aggregation functions in Section~\ref{subsubsec:aggregation_functions}.

For EPIC-Kitchens and EGTEA Gaze+, we use the feature provided by~\cite{furnari2019would} directly.
%For EPIC-Kitchens, we use the rgb feature provided by~\cite{furnari2019would}. For EGTEA Gaze+, following~\cite{furnari2019would}, we utilize TSN~\cite{wang2016temporal} model as the feature extractor in the observed information encoding step.
For 50 Salads, we simply use the feature provided by \cite{lea2017temporal}. For Breakfast, we use I3D~\cite{carreira2017quo} to extract the feature representation which will be released along with our source code.

\subsubsection{Training Details}
 
For EPIC-Kitchens and EGTEA Gaze+, when we train our model, we first randomly sample a training instance with 14~($o$ + $a$) frames before the target activity. Then, we split it into 8 training instances with different length of anticipation time-steps~(from 1 to 8). Finally, all instances are used to train our model jointly. We use SGD optimizer to train our model. The momentum and weight decay are set to 0.9 and 0.00005, respectively. The mini-batch size is 128. We also utilize dropout layer with dropout ratio 0.5. The provided action and object classes and the synthetic activity classes are utilized as labels to form $L_v$, $L_n$ and $L_{a}$. Specially, for EPIC-Kitchens, we set the initial learning rate as 0.1. The training procedure stops after 100 epochs. For the weight of each loss function, we set $\alpha$ as $0.01$ and $\beta$ as $0.8$, and the setting is determined via a cross-validation. For EGTEA Gaze+, the model is trained with an initial learning rate of 0.1 and 100 epochs. Through cross-validation, we choose $\alpha$ as $0.5$ and $\beta$ as $0.5$.

For 50 Salads and Breakfast, considering the dense anticipation protocol, we design a new training instance generation method. Specifically, we enlarge the values of $o$ and $a$ to $16$ and $16$. Besides, we use a temporal sliding window of 32~($o+a$) to generate training instances from the beginning to the end of each video. We use Adam optimizer to train our model. $\beta_1$, $\beta_2$ and weight decay are set to 0.9, 0.999 and 0.00005, respectively. We set the mini-batch size as 128. Dropout layer with dropout ratio 0.5 is also used. We utilize the activity labels to train $L_{a}$ and the mined action and object labels to train $L_v$ and $L_n$. For 50 Salads, the learning rate starts from 0.001. The training procedure stops after 100 epochs. We set $\alpha$ as $0.9$ and $\beta$ as $0.1$ via cross-validation for the weights of each loss function. For Breakfast, the model is trained with an initial learning rate of 0.01. The training procedure stops after 80 epochs. We choose $\alpha$ as $0.5$ and $\beta$ as $0.5$ through cross-validation.

All experiments are implemented under the pytorch framework. For datasets EGTEA Gaze+, Breakfast and 50 Salads, several activity categories contain multiple objects. The annotation template for most of these activity categories is `put~(or place) one object to~(or into) another object'~({\it e.g.}, `put egg to plate'). For these activity categories, the first object indicates the major object in this activity. Hence, we simply utilize the first object as the object label.

\subsection{Ablation Study on EPIC-Kitchens}
\label{subsec:epic-abliation}

%As analyzed in Section~\ref{sec:introduction}, there are three core elements we need to consider for getting precise anticipation results. 

To analyze the validity of each element in our proposed method, we carry out extensive ablation studies on EPIC-Kitchens and the results are shown in Table~\ref{table:epic-ablation-rgb}. 

%The detailed settings and results analysis are shown below.

\subsubsection{Baseline}
\label{subsubsec:baseline}
%\textbf{Baseline.}
For the baseline model (`Baseline' in Table \ref{table:epic-ablation-rgb}), the observed information encoding step is the same as SRL. After that, only a single $GRU$ layer is utilized in the recursive sequence prediction process to  predict the feature representation recursively at each anticipation time. Finally, a fully connected layer with softmax activation function is used to predict the target activity. As shown in Table~\ref{table:epic-ablation-rgb}, the top-5 accuracy of the baseline is 29.18\% at anticipation time 1s.

\begin{table*}[h]
	\small
	\renewcommand{\arraystretch}{1.05}
	\setlength{\tabcolsep}{2mm}
	\centering
	\caption{Ablation studies on EPIC-Kitchens. Given the baseline model, we explore the validity of each component.}
	\begin{tabular}{l|c|c|c|c|c|c|c|c|c|c|c}
		\hline
		\multirow{2}{*}{Setting} & \multirow{2}{*}{Revision} & \multirow{2}{*}{Reattend} & \multirow{2}{*}{Semantic Context} & \multicolumn{8}{c}{Top-5 Accuracy \% at different $\tau_a$ (s)} \\
		\cmidrule{5-12}
		& & & & 2 & 1.75 & 1.5 & 1.25 & 1.0 & 0.75 & 0.5 & 0.25 \\
		\hline 
		Baseline &   &   &  & 24.32  & 25.06  & 26.29  & 27.39  & 29.18  & 30.45  & 31.42  & 33.75 \\
		+Rev & \checkmark  &   &   & 24.42  & 25.82  & 27.76  & 28.64  &  30.27 & 31.13  & 32.68  & 34.84  \\
		+Rea &   & \checkmark  &   &  25.44 & 26.99  & 28.22  & 29.22  & 30.71  & 32.30  & 33.41 & 35.30  \\
		+SecCon &   &   & \checkmark  & 25.46  & 27.11  & 27.43  & 28.96  & 30.39  &  31.60 & 32.64  & 34.65  \\
		+Rev \& Rea & \checkmark & \checkmark &  & 25.56  & 26.81  & 28.24  &  29.32 & 31.23  & 32.58  & 33.75  & 35.32  \\
		+Rev \& SecCon &  \checkmark &   & \checkmark  & 25.58  & 26.77  & 27.76  & 28.96  & 30.89  & 32.20  & 33.67  &  34.98 \\
		+Rea \& SecCon &   & \checkmark & \checkmark  & 25.24  & 26.29  & 27.90  & 28.74 & 30.77  & 31.94  & 33.39  & 35.38  \\
		\hline
		\textbf{SRL} & \checkmark  & \checkmark  &  \checkmark  & \textbf{25.82} & \textbf{27.21} &  \textbf{28.52} &  \textbf{29.81} & \textbf{31.68} & \textbf{33.11} & \textbf{34.75}  & \textbf{36.89}  \\
		% RU &   &   &  & 25.44 & 26.89 & 28.32 & 29.42 & 30.83 & 32.00 & 33.31 & 34.47  \\
		\hline
	\end{tabular}
	\label{table:epic-ablation-rgb}
	%\vspace{-2ex}
\end{table*}

\begin{table*}[h]
	\small
	\renewcommand{\arraystretch}{1.05}
	\setlength{\tabcolsep}{3.5mm}
	\centering
	\caption{Ablation studies on the sampling methods and the number of samples $N$ on EPIC-Kitchens.}
	\begin{tabular}{l|l|c|c|c|c|c|c|c|c}
		\hline
		\multicolumn{2}{c|}{Setting} & \multicolumn{8}{c}{Top-5 Accuracy \% at different $\tau_a$ (s)} \\
		\cmidrule{1-10}
		Sampling Method & $N$ & 2 & 1.75 & 1.5 & 1.25 & 1.0 & 0.75 & 0.5 & 0.25 \\
		%\cmidrule{2-9}
		\hline
		same video &  128 & 25.04 & 26.09 & 27.53 & 28.56 & 30.15 & 31.05 & 32.20 & 34.21 \\
		other video & 128 & 25.20 & 26.47 & 27.88 & 28.34 & 30.37 & 31.30 & 32.92 & 34.77\\
		all video & 128 & 25.26 & 26.83 & 28.08 & 28.94 & 30.31 & 31.74 & 33.23 & 34.39 \\
		all video & 32 & 25.12 & 26.01 & 27.90 & 28.14 & 30.15 & 31.92 & 32.76 & 35.06 \\
		all video & 64 & 25.66 & 27.01 & 28.06 & 28.90 & 30.33 & 31.82 & 33.21 & 35.40 \\
		all video & 256 & 25.42 & 26.85 & 27.61 & 28.14 & 30.19 & 31.66 & 33.15 & 34.84 \\
		\hline
	\end{tabular}
	\label{table:sample-number-method}
	%\vspace{-2ex}
\end{table*}

\subsubsection{Baseline+Rev}
\label{subsubsec:baseline_rev}
%\textbf{Baseline+Rev.} 
As shown in Table~\ref{table:epic-ablation-rgb}, compared to baseline model, the representation revision operation~(`+Rev') boosts the activity anticipation performance from 29.18\% to 30.27\% at anticipation time 1s, which proves the effectiveness of the representation revision. Actually, at all anticipation times~({\it i.e.}, 0.25s to 2s), the representation revision operation can lead to performance improvements. By comparing the row of `+Rea \& SecCon' and row of `SRL' in Table~\ref{table:epic-ablation-rgb}, we can see that after removing this operation, the performance is degraded to varying degrees at all anticipation times, which further proves the validity of the representation revision operation.
The representation revision operation demonstrates even larger improvement with shorter time windows. For example, the performance gap between `Baseline' and `Baseline+Rev' at anticipation time 0.5s is 1.26\% that is larger than 0.76\% at 1.75s. This is mainly because that the feature representation becomes more difficult to be revised as anticipation time increases. Although our model can alleviate the error accumulation, it will inevitably cause a certain degree of error accumulation. Thus, for longer anticipation time, the representation revision operation will face greater challenge compared to situation of shorter anticipation time. 

%The performance improvement is not as good as that of shorter anticipation time.

The core of the representation revision operation is the contrastive loss. We sample negative samples randomly from video clips with different activity labels as the selected negative set. We set the value of $N$ as $128$. Moreover, we conduct experiments to verify the impact of the samping methods and the number of samples $N$, as shown in Table~\ref{table:sample-number-method}. `same video' means we sample negative samples randomly from video clips that have the same video id as the positive sample. `other video' means we sample negative samples randomly from video clips that have different video id from the positive sample. `all video' means we sample negative samples randomly from all video clips of the training set. 

From the Table~\ref{table:sample-number-method}, we can get several important observations. First, the `all video' sampling method achieves better results at most anticipation time-steps. However, compared with other sampling methods, the performance advantage is not obvious. On EPIC-Kitchens, videos are captured by different actors in the kitchen scenes. These videos contain diversified and similar semantic information, which leads to more hard negatives in the sampled batch. Hence, sampling from all video clips can better guarantee the diversity and the similarity of the negative samples compared to positive sample, which helps to get more accurate anticipation results. Accordingly, we choose this sampling method in our experiments. Second, the number of samples has a slight influence on our model. Different number of samples has their own prediction performance advantages at some anticipation times. Hence, we choose the appropriate number of samples according to the convenience of the implementation.

%here

\begin{table*}[h]
	\small
	\renewcommand{\arraystretch}{1.05}
	\setlength{\tabcolsep}{3.5mm}
	\centering
	\caption{Egocentric activity anticipation results on the EPIC-Kitchens with different modality features.}
	\begin{tabular}{l|l|c|c|c|c|c|c|c|c}
		\hline
		\multicolumn{2}{c|}{Setting} & \multicolumn{8}{c}{Top-5 Accuracy \% at different $\tau_a$ (s)} \\
		\cmidrule{1-10}
		Mode & Model & 2 & 1.75 & 1.5 & 1.25 & 1.0 & 0.75 & 0.5 & 0.25 \\
		\hline
		RGB & RU \cite{furnari2019would} & 25.44 & 26.89 & 28.32 & 29.42 & 30.83 & 32.00 & 33.31 & 34.47  \\
		& \textbf{SRL} & \textbf{25.82} & \textbf{27.21} & \textbf{28.52} & \textbf{29.81} & \textbf{31.68} & \textbf{33.11} & \textbf{34.75} & \textbf{36.89} \\
		\hline
		FLOW & RU \cite{furnari2019would} & 17.38 & 18.04 & 18.91 & 19.97 & 21.42 & 22.37 & 23.49 & 24.18  \\
		& \textbf{SR}L & \textbf{17.84} & \textbf{18.85} & \textbf{19.85} & \textbf{20.94} & \textbf{21.72} & \textbf{23.23} & \textbf{24.62} & \textbf{25.78} \\
		\hline
		OBJ & RU \cite{furnari2019would} & 24.54 & 25.58 & 26.63 & 28.32 & 29.89 & 30.85 & 31.82 & 33.39  \\
		& \textbf{SRL} & \textbf{25.32} & \textbf{26.59} & \textbf{27.47} & \textbf{28.56} & \textbf{30.15} & \textbf{31.23} & \textbf{33.09} & \textbf{34.53} \\
		\hline
		RGB + OBJ & SRL & 29.95 & 31.19 & 32.62 & 34.01 & 35.32 & 36.56 & 38.46 & 40.12 \\
		%\hline 
		RGB + FLOW & SRL & 26.99 & 28.06 & 29.20 & 30.73 & 31.94 & 33.37 & 35.26 & 37.47 \\    
		%\hline 
		OBJ + FLOW & SRL & 26.93 & 28.16 & 29.06 & 30.43 & 32.10 & 33.17 & 34.59 & 36.50 \\        
		\hline
		Late Fusion & RU(Late) \cite{furnari2019would} & 29.10 & 29.77 & 31.72 & 33.09 & 34.23 & 35.28 & 36.10 & 37.61  \\
		& \textbf{SRL(Late)} & \textbf{29.83} & \textbf{31.07} & \textbf{31.92} & \textbf{33.77} & \textbf{35.36} & \textbf{36.63} & \textbf{38.56} & \textbf{40.43} \\
		\hline
		Attention Fusion & RU(Atten) \cite{furnari2019would} & 29.49 & 30.75 & 32.24 & 33.41 & 35.34 & 36.34 & 37.37 & 39.00  \\
		& \textbf{SRL(Atten)} & \textbf{30.15} & \textbf{31.28} & \textbf{32.36} & \textbf{34.05} & \textbf{35.52} & \textbf{36.77} & \textbf{38.60} & \textbf{40.49} \\	     
		\hline
	\end{tabular}
	\label{table:epic-multiple-mode}
	%\vspace{-2ex}
\end{table*}

\subsubsection{Baseline+Rea}
\label{subsubsec:baseline_rea}
%\textbf{Baseline+Rea.}
From Table \ref{table:epic-ablation-rgb}, with the dynamically reattending operation~(`+Rea' in the table), the performance is 1.53\% higher than baseline at anticipation time 1s. The performance improvements are also achieved at other anticipation times, showing that this operation is useful for activity anticipation. Its effectiveness can be further demonstrated by comparing the results in the line `+Rev \& SecCon' and line `SRL' of the Table~\ref{table:epic-ablation-rgb}. Besides, by seeing the performance gap between `Baseline' and `Baseline+Rea' at anticipation time 0.5s~(1.99\%) and 1.75s~(1.93\%), the dynamically reattending operation gives similar improvement for different anticipation times. This phenomenon is different from the representation revision operation. This is mainly because that the dynamically reattending operation can obtain useful information at different anticipation times. At each anticipation time, even though the generated feature representation contains noise, it can still coarsely represent the current video content. Accordingly, the dynamically reattending operation can use the predicted representation to capture useful observed information to some extent. Therefore, the dynamically attending operation is less sensitive to the anticipation time.

\subsubsection{Baseline+SecCon}
\label{subsubsec:baseline_seccon}
%\textbf{Baseline+SecCon.}
In Table \ref{table:epic-ablation-rgb}, by exploring semantic context information related to the target activity (`+SecCon'), the performance is 1.21\% higher than baseline. Moreover, by comparing the results of the line `+Rev \& Rea' and the line `SRL', we can find the performance is degraded at each anticipation time by removing the semantic context exploration operation. These phenomenons verify the effectiveness of the semantic context exploration operation. 

%\textcolor{blue}
To show the validity of this operation more clearly, we visualize the semantic context information predicted by SRL in Figure~\ref{fig:model-visualize}. Take the first one as an example, at the target anticipation time, our model obtains the action~(`roll') and object~(`dough'). Indeed, the `roll' reveals the action of the target activity and the `dough' reveals the object involved in the target activity. The obtained action and object do closely relate to the target activity. Hence, the semantic context helps us get more accurate anticipation results. Similar conclusions can also be drawn from other examples.

\subsubsection{Baseline+Combination of Any Two Components}
\label{subsubsec:baseline_anytwo}
%\textbf{Baseline + combination of any two.}
The validity of each component of SRL has been demonstrated in the above experiments. We also explore the effectiveness of any two combination of these components, the results are shown in the 5th\textasciitilde7th row in Table~\ref{table:epic-ablation-rgb}. We can find that the performance of the combination of two components is higher than that of a single component. For example, the 5th row of Table~\ref{table:epic-ablation-rgb} shows the result of the combination of the representation revision operation and dynamically attending operation, the top-5 accuracy at all anticipation times is higher than any single operation. %The same is true for the other two combinations.

\subsubsection{On Combining Multiple Modalities}
\label{subsubsec:multiple_modalities}
%\textbf{On combining multiple modalities.}
So far, all experiments are based on RGB features. To further verify the validity of SRL, we conduct extensive experiments on other types of feature representations~({\it i.e.}, optical-flow and object features). The results are shown in Table~\ref{table:epic-multiple-mode}. `RU(Late)' and `SRL(Late)' indicate the models using late fusion strategy to merge the prediction results of the three feature modalities. `RU(Atten)' and `SRL(Atten)' indicate the models using an attention module to combine the results of the three feature modalities.

For a fair comparison, we use the pre-computed optical-flow and object features provided by~\cite{furnari2019would}. The optical-flow features are extracted using a Batch Normalized Inception CNN. The object features are extracted using Faster R-CNN~\cite{girshick2015fast}. See~\cite{furnari2019would} for more details about the utilized features. For models that use optical-flow or object features, the observed time-step $o$ and the anticipation time-step $a$ are also set to 6 and 8, respectively. A simple GRU layer is used as the aggregation function $\Phi$. We use SGD optimizer with the mini-batch size of 128. For model that uses optical-flow features, the initial learning rate is set as 0.05. The momentum and weight decay are set to 0.9 and 0.00005, respectively. The training procedure stops after 100 epochs. For the weights of each loss function, we set $\alpha$ as $0.5$ and $\beta$ as $0.5$, which is determined via cross-validation. For model that uses object features, we train the model with an initial learning rate of 0.1. The momentum and weight decay are set to 0.9 and 0.0001, respectively. The training procedure stops after 100 epochs. Through cross-validation, we set $\alpha$ as $0.8$ and $\beta$ as $0.8$ for the loss function.

\begin{table*}[h]
	\small
	\renewcommand{\arraystretch}{1.05}
	\setlength{\tabcolsep}{1.5mm}
	\centering
	\caption{Results on the EPIC-Kitchens in terms of top-5 accuracy at different anticipation time-steps. `Act.' means activity.}
	\begin{tabular}{l|c|c|c|c|c|c|c|c|c|c|c|c|c|c}
		\hline
		\multirow{2}{*}{Model} & \multicolumn{8}{c|}{Top-5 Accuracy \% at different $\tau_a$ (s)} & \multicolumn{3}{c|}{Top-5 Acc. \% @ 1s} & \multicolumn{3}{c}{M Top-5 Rec. \% @ 1s}\\
		\cmidrule{2-15}
		& 2 & 1.75 & 1.5 & 1.25 & 1.0 & 0.75 & 0.5 & 0.25 & Action & Object & Act. & Action & Object & Act. \\
		\hline 
		DMR \cite{vondrick2016anticipating} & / & / & / & / & 16.86 & / & / & / & 73.66 & 29.99 & 16.86 & 24.50 & 20.89 & 03.23 \\
		ATSN \cite{damen2018scaling} & / & / & / & / & 16.29 & / & / & / & 77.30 & 39.93 & 16.29 & 33.08 & 32.77 & 07.60 \\
		MCE \cite{furnari2018leveraging} & / & / & / & / & 26.11 & / & / & / & 73.35 & 38.86 & 26.11 & 34.62 & 32.59 & 06.50 \\
		VN-CE \cite{damen2018scaling} & / & / & / & / & 17.31 & / & / & / & 77.67 & 39.50 & 17.31 & 34.05 & 34.50 & 07.73 \\
		%TE-TOP3 \cite{lapin2017analysis} & / & / & / & / & 25.25 & / & / & / & 73.53 & 39.54 & 25.25 & 44.14 & 36.69 & 06.14 \\
		%TE-TOP5 \cite{lapin2017analysis} & / & / & / & / & 25.74 & / & / & / & 73.75 & 40.10 & 25.74 & 43.77 & 36.38 & 05.79 \\
		SVM-TOP3 \cite{berrada2018smooth} & / & / & / & / & 25.42 & / & / & / & 72.70 & 38.41 & 25.42 & 41.90 & 34.69 & 5.32 \\
		SVM-TOP5 \cite{berrada2018smooth} & / & / & / & / & 24.46 & / & / & / & 69.17 & 36.66 & 24.46 & 40.27 & 32.69 & 05.23 \\
		VNMCE+T3 \cite{furnari2018leveraging} & / & / & / & / & 25.95 & / & / & / & 74.05 & 39.18 & 25.95 & 40.17 & 34.15 & 05.57 \\
		VNMCE+T5 \cite{furnari2018leveraging} & / & / & / & / & 26.01 & / & / & / & 74.07 & 39.10 & 26.01 & 41.62 & 35.49 & 05.78 \\
		ED \cite{gao2017red} & 21.53 & 22.22 & 23.20 & 24.78 & 25.75 & 26.69 & 27.66 & 29.74 & 75.46 & 42.96 & 25.75 & 41.77 & 42.59 & 10.97 \\
		FN \cite{de2018modeling} & 23.47 & 24.07 & 24.68 & 25.66 & 26.27 & 26.87 & 27.88 & 28.96 & 74.84 & 40.87 & 26.27 & 35.30 & 37.77 & 06.64 \\
		RL \cite{ma2016learning} & \textbf{25.95} & 26.49 & 27.15 & 28.48 & 29.61 & 30.81 & 31.86 & 32.84 & 76.79 & 44.53 & 29.61 & 40.80 & 40.87 & 10.64 \\
		EL \cite{jain2016recurrent} & 24.68 & 25.68 & 26.41 & 27.35 & 28.56 & 30.27 & 31.50 & 33.55 & 75.66 & 43.72 & 28.56 & 38.70 & 40.32 & 08.62 \\
		%RU \cite{furnari2019would} & 29.49 & 30.75 & 32.24 & 33.41 & 35.34 & 36.34 & 37.37 & 39.00 & 79.57 & 51.79 & 35.34 & 43.82 & 49.90 & 15.11 \\
		RU-RGB \cite{furnari2019would} & 25.44 & 26.89 & 28.32 & 29.42 & 30.83 & 32.00 & 33.31 & 34.47 & / & / & 30.83 & / & / & / \\
		\hline
		\textbf{SRL} & 25.82 & \textbf{27.21} & \textbf{28.52} & \textbf{29.81} & \textbf{31.68} & \textbf{33.11} & \textbf{34.75} & \textbf{36.89} & \textbf{78.90} & \textbf{47.65} & \textbf{31.68} & \textbf{42.83} & \textbf{47.64} & \textbf{13.24}\\
		\hline
	\end{tabular}
	\label{table:epic-results}
	%\vspace{-2ex}
\end{table*}

It can be seen from Table~\ref{table:epic-multiple-mode} that our model achieves better anticipation performance under different feature modalities~({\it i.e.}, RGB, OBJ or FLOW) at all anticipation times compared to~\cite{furnari2019would}. Compared to the results using single feature, our model can also achieve higher anticipation accuracy at all anticipation times under the setting of any two feature modalities, {\it i.e.}, RGB+OBJ, RGB+FLOW, and OBJ+FLOW. This phenomenon suggests that the OBJ and FLOW features are helpful for getting more accurate anticipation results. By comparing the results of RU(Late) and SRL(Late), we can find that our SRL(Late) achieves better performance at all anticipation times using the same features and fusion method. Specifically, at anticipation time 0.25s, our model can improve the top-5 accuracy from 37.61\% to 40.21\%, resulting in a 2.6\% increase. Actually, the abundant visual information and semantic context information contained in the video content is not fully utilized by RU(Late). The better performance of SRL demonstrates the effectiveness and necessity of each component in our model.

Moreover, the performance of SRL(Late) is comparable to that of RU(Atten) at all anticipation time stamps. Note that RU(Atten) designs a Modality ATTention (MATT) module that calculates attention scores to indicate the relative importance of each feature modality for the final anticipation. Therefore, we design a similar attention fusion method to fuse the results of different feature modalities. At the target anticipation time-step, we first concatenate the observed video clip representations of each feature modality. Then, we use an MLP network with three layers to produce the modality-wise attention score. Finally, the attention score is used to fuse the anticipation results of each feature modality. From the experimental results, we can see that SRL(Atten) can obtain higher performance at all anticipation time stamps compared to RU(Atten) and SRL(Late). In summary, the above experiments verify the validity of single feature and feature combination. For simplicity, in the following experiments, we only use the RGB features in our model. 

%here
\subsubsection{Different Aggregation Functions}
\label{subsubsec:aggregation_functions}
In our model, we consider three crucial factors when we choose GRU as our aggregation function~$\Phi$. First, as a sequence model, GRU can encode the observed video information more effectively compared to pooling methods. The aggregated representation $h_o$ at the last observed time-step contains complex history information about the observed video clip. Second, adjacent video frames are more likely to have strong correlation. Using GRU, the aggregated representation $h_o$ can be pushed to the feature representation at the last observation time-step. It assists us to get more accurate prediction results at the first anticipation time-step, and benefit consequent anticipations. Third, compared to LSTM, GRU has fewer parameters. We also try other aggregation functions to compare the experimental results. They are average pooling~(Avg), max pooling~(Max) and LSTM~(LSTM). The results are shown in Table~\ref{table:aggregation-function}. We can find that GRU can get better results compared to other aggregation functions. %%Hence, we choose the GRU.

\subsection{Experiments on Egocentric Video}

\subsubsection{EPIC-Kitchens}
We compare SRL with the following anticipation models: DMR~\cite{vondrick2016anticipating}, ATSN~\cite{damen2018scaling}, MCE~\cite{furnari2018leveraging}, VN-CE~\cite{damen2018scaling}, SVM-TOP3~\cite{berrada2018smooth}, SVM-TOP5~\cite{berrada2018smooth}, VNMCE+T3~\cite{furnari2018leveraging}, VNMCE+T5~\cite{furnari2018leveraging}, ED~\cite{gao2017red}, FN~\cite{de2018modeling}, RL~\cite{ma2016learning}, EL~\cite{jain2016recurrent} and RU-RGB~\cite{furnari2019would}. Note that we only compare RU using RGB features~(RU-RGB), and the results using other features are shown in Table~\ref{table:epic-multiple-mode}. We use Top-5 accuracy for activity prediction at different anticipation times~({\it i.e.}, 0.25s\textasciitilde2s), Top-5 accuracy and mean Top-5 recall for action, object and activity prediction at anticipation time 1s to evaluate performance. The experimental results are shown in Table~\ref{table:epic-results}. 

Table~\ref{table:epic-results} clearly shows that SRL achieves the start-of-the-art anticipation performance at most anticipation times. ATSN model, which simply uses the recognition model TSN~\cite{wang2016temporal} for activity prediction, only achieves 16.86\% Top-5 accuracy at anticipation time 1s. The low performance suggests that we need to design special models to adapt to the video data property of the activity anticipation task. Methods like MCE, FN, RL and EL anticipate the target activity from the observed video clip directly. Instead, methods like RU-RGB and SRL are developed upon the recursive anticipation framework. By comparing the performance differences between these two types of methods, we can find that the recursive anticipation pattern is more suitable for activity anticipation task. 

\begin{table}[t]
	\small
	\renewcommand{\arraystretch}{1.05}
	\setlength{\tabcolsep}{1mm}
	\centering
	\caption{Ablation studies about aggregation function on EPIC-Kitchens.}
	\begin{tabular}{l|c|c|c|c|c|c|c|c}
		\hline
		\multirow{2}{*}{Setting} & \multicolumn{8}{c}{Top-5 Accuracy \% at different $\tau_a$ (s)} \\
		\cmidrule{2-9}
		& 2 & 1.75 & 1.5 & 1.25 & 1.0 & 0.75 & 0.5 & 0.25 \\
		\hline
		Avg & \textbf{25.93} & 27.21 & \textbf{28.86} & 29.67 & 31.03 & 32.10 & 33.02 & 34.59 \\
		Max & 26.31 & 27.13 & 27.84 & 29.10 & 30.61 & 31.82 & 32.90 & 34.13\\
		LSTM & 24.96 & 26.23 & 27.63 & 29.00 & 30.79 & 31.84 & 32.98 & 35.04 \\
		GRU & 25.82 & \textbf{27.21} & 28.52 & \textbf{29.81} & \textbf{31.68} & \textbf{33.11} & \textbf{34.75} & \textbf{36.89} \\
		\hline 
	\end{tabular}
	\label{table:aggregation-function}
	%\vspace{-2ex}
\end{table}

\begin{table*}[h]
	\small
	\renewcommand{\arraystretch}{1.05}
	\setlength{\tabcolsep}{1.2mm}
	\centering
	\caption{Results on the EPIC-Kitchens test set with seen (\textbf{S1}) and unseen (\textbf{S2}) kitchens.}
	\begin{tabular}{c|l|c|c|c|c|c|c|c|c|c|c|c|c}
		\hline
		\multirow{2}{*}{Setting} & \multirow{2}{*}{Model} & \multicolumn{3}{c|}{Top-1 Acc. \% @ 1s} & \multicolumn{3}{c|}{Top-5 Acc. \% @ 1s} & \multicolumn{3}{c|}{Avg Class Precision. \% @ 1s} & \multicolumn{3}{c}{Avg Class Recall. \% @ 1s}\\
		\cmidrule{3-14}
		& & Action & Object & Act. & Action & Object & Act. & Action & Object & Act. & Action & Object & Act.\\
		\hline %\hline 
		\multirow{5}{*}{\textbf{S1}}
		& 2SCNN \cite{damen2018scaling} & 29.76 & 15.15 & 04.32 & 76.03 & 38.56 & 15.21 & 13.76 & 17.19 & 02.48 & 07.32 & 10.72 & 01.81\\
		& ATSN \cite{damen2018scaling} & 31.81 & 16.22 & 06.00 & 76.56 & 42.15 & 28.21 & 23.91 & 19.13 & 03.13 & 09.33 & 11.93 & 02.39 \\
		& MCE \cite{furnari2018leveraging} & 27.92 & 16.09 & 10.76 & 73.59 & 39.32 & 25.28 & 23.43 & 17.53 & 06.05 & 14.79 & 11.65 & 05.11 \\
		& RU \cite{furnari2019would} & 33.04 & 22.78 & 14.39 & 79.55 & 50.95 & 33.73 & 25.50 & 24.12 & 07.37 & \textbf{15.73} & 19.81 & 07.66 \\
		& TAR \cite{sener2020temporal} & \textbf{37.87} & \textbf{24.10} & \textbf{16.64} & \textbf{79.74} & \textbf{53.98} & \textbf{36.06} & \textbf{36.41} & 25.20 & \textbf{09.64} & 15.67 & \textbf{22.01} & \textbf{10.05} \\
		\cmidrule{2-14}
		& \textbf{SRL} & 34.89 & 22.84 & 14.24 & 79.59 & 52.03 & 34.61 & 28.29 & \textbf{25.69} & 06.45 & 12.19 & 19.16 & 06.34 \\
		\hline %\hline
		\multirow{5}{*}{\textbf{S2}}
		& 2SCNN \cite{damen2018scaling} & 25.23 & 09.97 & 02.29 & 68.66 & 27.38 & 09.35 & 16.37 & 06.98 & 00.85 & 05.80 & 06.37 & 01.14 \\
		& ATSN \cite{damen2018scaling} & 25.30 & 10.41 & 02.39 & 68.32 & 29.50 & 06.63 & 07.63 & 08.79 & 00.80 & 06.06 & 06.74 & 01.07 \\
		& MCE \cite{furnari2018leveraging} & 21.27 & 09.90 & 05.57 & 63.33 & 25.50 & 15.71 & 10.02 & 06.88 & 01.99 & 07.68 & 06.61 & 02.39 \\
		& RU \cite{furnari2019would} & 27.01 & 15.19 & 08.16 & 69.55 & 34.38 & 21.10 & 13.69 & 09.87 & 03.64 & \textbf{09.21} & 11.97 & 04.83 \\
		& TAR \cite{sener2020temporal} & \textbf{29.50} & \textbf{16.52} & \textbf{10.04} & 70.13 & \textbf{37.83} & \textbf{23.42} & \textbf{20.43} & \textbf{12.95} & \textbf{04.92} & 08.03 & \textbf{12.84} & \textbf{06.26} \\
		\cmidrule{2-14}
		& \textbf{SRL} & 27.42 & 15.47 & 08.88 & \textbf{71.90} & 36.80 & 22.06 & 20.23 & 12.48 & 02.84 & 07.83 & 12.25 & 04.33 \\
		\hline
	\end{tabular}
	\label{table:epic-results-testset}
	%\vspace{-2ex}
\end{table*}

\begin{table*}[h]
	\small
	\renewcommand{\arraystretch}{1.05}
	\setlength{\tabcolsep}{6mm}
	\centering
	\caption{Egocentric activity anticipation results on EGTEA Gaze+.}
	\begin{tabular}{l|c|c|c|c|c|c|c|c}
		\hline
		\multirow{2}{*}{Model} & \multicolumn{8}{c}{Top-5 Accuracy \% at different $\tau_a$ (s)}\\
		\cmidrule{2-9}
		& 2 & 1.75 & 1.5 & 1.25 & 1.0 & 0.75 & 0.5 & 0.25 \\
		\hline %\hline 
		DMR \cite{vondrick2016anticipating} & / & / & / & / & 55.70 & / & / & / \\
		ATSN \cite{damen2018scaling} & / & / & / & / & 40.53 & / & / & / \\
		MCE \cite{furnari2018leveraging} & / & / & / & / & 56.29 & / & / & / \\
		ED \cite{gao2017red} & 45.03 & 46.22 & 46.86 & 48.36 & 50.22 & 51.86 & 49.99 & 49.17 \\
		FN \cite{de2018modeling} & 54.06 & 54.94 & 56.75 & 58.34 & 60.12 & 62.03 & 63.96 & 66.45 \\
		RL \cite{ma2016learning} & 55.18 & 56.31 & 58.22 & 60.35 & 62.56 & 64.65 & 67.35 & 70.42 \\
		EL \cite{jain2016recurrent} & 55.62 & 57.56 & 59.77 & 61.58 & 64.62 & 66.89 & 69.60 & 72.38  \\
		RU \cite{furnari2019would} & 56.82 & 59.13 & 61.42 & 63.53 & 66.40 & 68.41 & 71.84 & 74.28 \\
		\hline %\hline
		\textbf{SRL} & \textbf{59.69} & \textbf{61.79} & \textbf{64.93} & \textbf{66.45} & \textbf{70.67} & \textbf{73.49} & \textbf{78.02} & \textbf{82.61} \\
		\hline
	\end{tabular}
	\label{table:gtea-results}
	\vspace{-1ex}
\end{table*}

RU-RGB employs a sequence completion pre-training in their model to improve the anticipation performance. Without this pre-training setup, we can still achieve better performance.
In fact, there are no enhancements to the predicted feature representation in RU-RGB. The better performance of SRL verifies the necessary of exploiting the informative visual cues contained in the video in the anticipation stage. When inspecting the Top-5 accuracy and the mean Top-5 recall for action, object and activity prediction at anticipation time 1s, we can find a relatively large improvement compared to previous methods. The improvement of action and object prediction performance shows that our semantic context exploration operation is effective, and this indeed assists us to obtain better activity prediction results. %This finding can illustrate the validity of our SRL from another aspect. 

% We also conduct experiments on the EPIC-Kitchens test set with seen~(\textbf{S1}) and unseen~(\textbf{S2}) kitchens. The \textbf{S1} indicates the test set includes scenes appearing in the training set. The \textbf{S2} means the test set includes scenes not appearing in the training set. We use the official evaluation metrics, {\it i.e.}, Top-1 accuracy, Top-5 accuracy, Average Class Precision and Average Class Recall. 
We also conduct experiments on the EPIC-Kitchens test set with seen~(\textbf{S1}) and unseen~(\textbf{S2}) kitchens. The results are shown in Table~\ref{table:epic-results-testset}. We can find that SRL obtains better anticipation performance than existing methods except TAR at most evaluation metrics, especially on \textbf{S2}. Essentially, TAR creates ensembles of multi-scale feature representations from the observed video clip. This operation is beneficial to predict the next activity. Instead, our SRL addresses the error accumulation issue over long periods of anticipation time. The Top-1 accuracy and Top-5 accuracy metrics are micro-averaged while the Average Class Precision and Average Class Recall metrics are macro-averaged. In a multi-class classification task, the micro-average is preferable if there exists class imbalance. For EPIC-Kitchens, the distribution of categories is imbalance. Accordingly, the higher performance on Top-1 and Top-5 accuracy further verifies the effectiveness of our model.

%-------------
% ###########################################################################

\subsubsection{EGTEA Gaze+}
We compare SRL with other anticipation models on EGTEA Gaze+, including DMR~\cite{vondrick2016anticipating}, ATSN~\cite{damen2018scaling}, MCE~\cite{furnari2018leveraging}, ED~\cite{gao2017red}, FN~\cite{de2018modeling}, RL~\cite{ma2016learning}. EL~\cite{jain2016recurrent} and RU~\cite{furnari2019would}. We evaluate the performance using Top-5 accuracy at different anticipation times~({\it i.e.}, 0.25s\textasciitilde2s). The comparison results are shown in Table~\ref{table:gtea-results}. We can find that the performance of SRL is better than all competitors at all anticipation times.

\begin{figure*}[h]
	\begin{center}
		\includegraphics[width=0.9\linewidth]{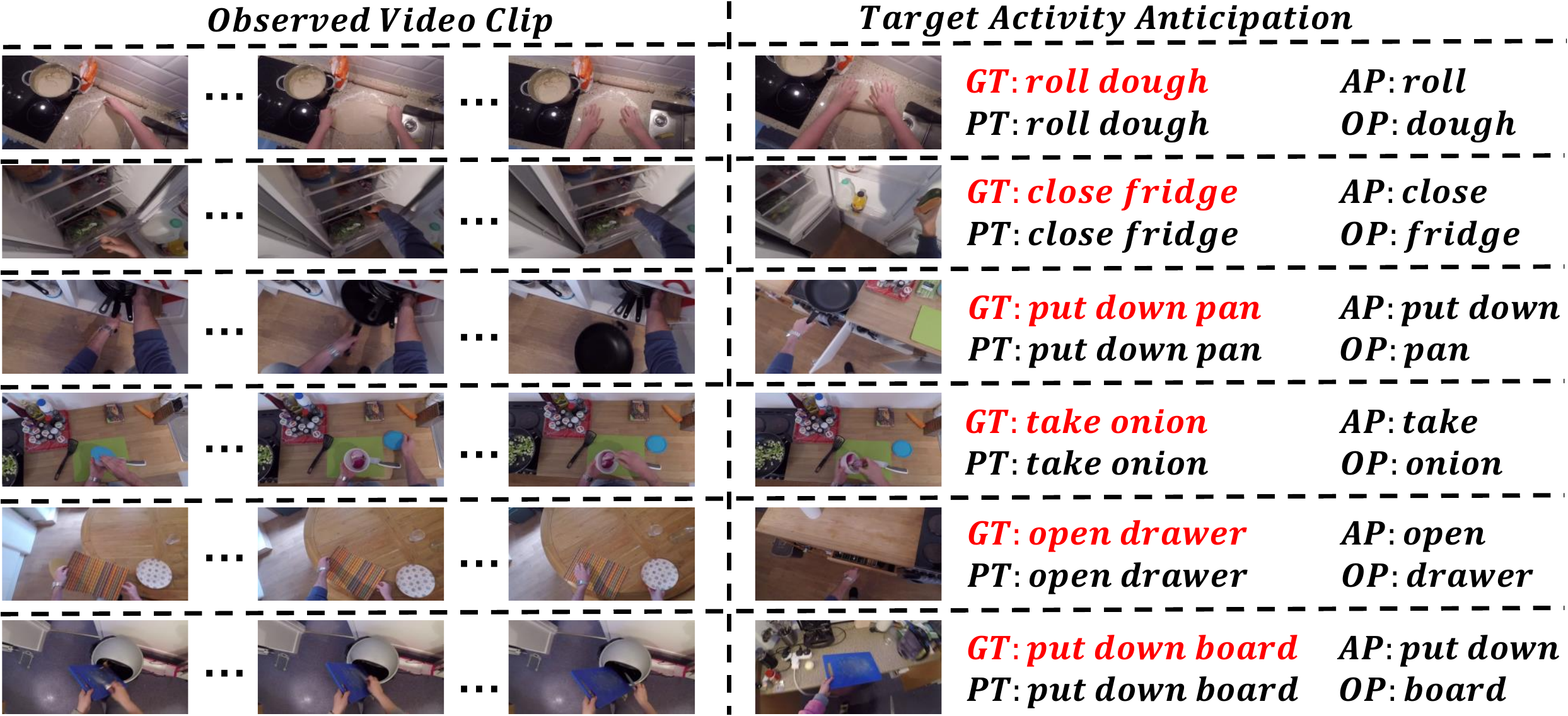}
	\end{center}
	\caption{The anticipation result visualization. In each example, the observed video clips are shown on the left. The target activity frame and its ground-truth activity category~(marked in red) and the predicted activity, action, object category are shown on the right. `GT' means ground-truth, `PT' means activity prediction, `AP' means action prediction and `OP' means object prediction.}
	\label{fig:model-visualize}
	%\vspace{-2ex}
\end{figure*}

\begin{figure*}[ht]
	\begin{center}
		\includegraphics[width=0.9\linewidth]{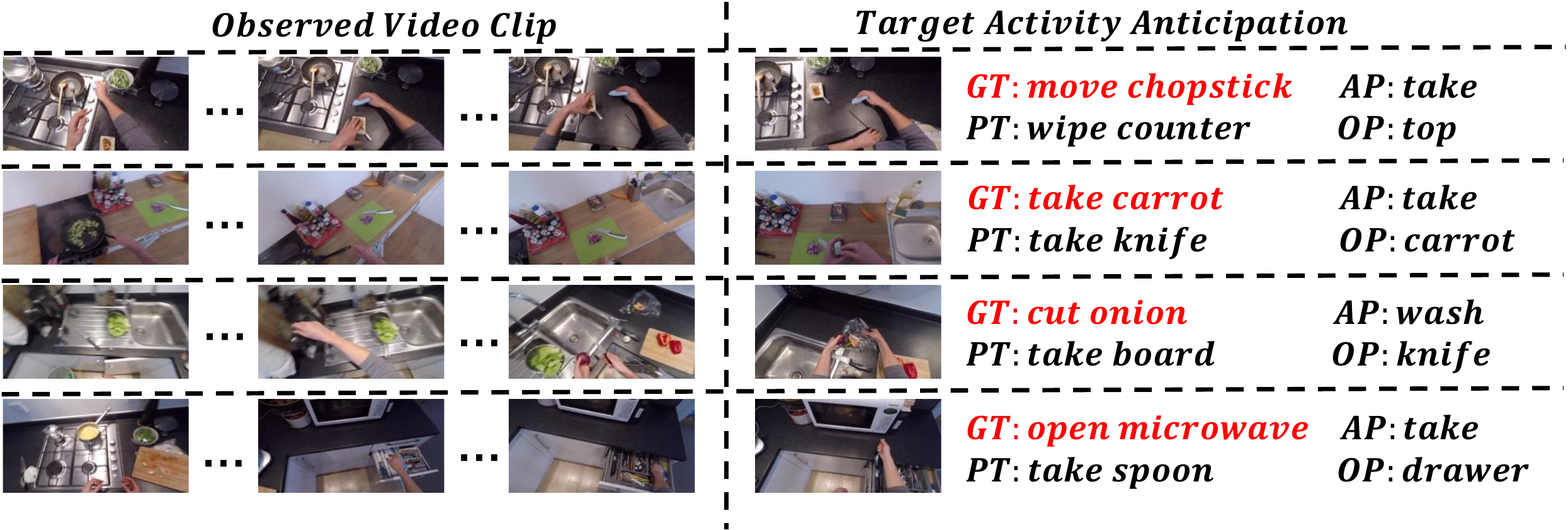}
	\end{center}
	\caption{The failure case visualization. In each example, the observed video frames are shown on the left. The target activity frame and its ground-truth activity  category~(mark in red) and the predicted activity, action and object category are shown on the right. `GT' means ground-truth, `PT' means activity  prediction, `AP' means action prediction and `OP' means object prediction.}
	\label{fig:model-visualize-failure}
	%\vspace{-2ex}
\end{figure*}

\subsection{Qualitative Analysis}
\label{subsection:visualization}

% In each instance, the observed video clip is shown on the left. The target anticipation video frame, ground-truth activity category~(mark in red) and predicted activity category are shown on the right. 

To show the anticipation capability of SRL more clearly, we visualize the anticipation results on EPIC-Kitchens in Figure~\ref{fig:model-visualize}. Take the first one as an example. The length of the observed video clip is 1.5s, we can see that the `dough' is placed on the `cutting board' step by step and the related `roll pin' can also be seen in the video. After watching this video, our model can correctly predict the next activity `rolling dough'. Moreover, in the last example, our model effectively models the information contained in the observed video and accurately predicts the next activity `put down board'. From the above examples, it can be seen that our SRL can make good use of the observed information and produce accurate prediction.

In order to have a deeper understanding of our model, we also visualize some failure cases in Figure~\ref{fig:model-visualize-failure}. Take the first one as an example, the ground-truth activity is `move chopstick' and our prediction is `wipe counter'. From the whole video, we can find that the `move chopstick' and the `wipe counter' are two consecutive activities. In the observed video clip, our model can see the activity `wipe counter'. In the target anticipation time-step, key objects `chopstick' and `top' both appear in the scene. Unfortunately, our model predicts the target activity as `wipe counter' and captures the `top' as key objects. As we can see from the target frame, the person is moving chopstick with one hand and wiping counter with the other. There is some overlap between the two consecutive activities. Even for activity recognition, this is also a hard example to distinguish. Hence, in this case the anticipation model fails.

\begin{table*}[ht]
	\small
	\renewcommand{\arraystretch}{1.05}
	\setlength{\tabcolsep}{1.1mm}
	\centering
	\caption{Third-person activity anticipation results on 50 Salads and Breakfast. RU-RGB* means our reimplementation of RU\cite{furnari2019would} using RGB features.}
	\begin{tabular}{l|c|c|c|c|c|c|c|c|c|c|c|c|c|c|c|c}
		\hline
		Dataset & \multicolumn{8}{c|}{50 Salads}  & \multicolumn{8}{c}{Breakfast} \\
		\hline
		Observed & \multicolumn{4}{c|}{20 \%} & \multicolumn{4}{c|}{30 \%}  & \multicolumn{4}{c|}{20 \%} & \multicolumn{4}{c}{30 \%} \\
		\hline
		Predicted & 10 \% & 20 \% & 30 \% & 50 \% & 10 \% & 20 \% & 30 \% & 50 \% & 10 \% & 20 \% & 30 \% & 50 \% & 10 \% & 20 \% & 30 \% & 50 \% \\
		\hline %\hline
		Nearest-Neighbor & 19.04 & 16.10 & 14.13 & 10.37 & 21.63 & 15.48 & 13.47 & 13.90  & 16.42 & 15.01 & 14.47 & 13.29 & 19.88 & 18.64 & 17.97 & 16.57 \\
		%\textcolor{blue}{RU-RGB* \cite{furnari2019would}} & \textcolor{blue}{22.21} & \textcolor{blue}{17.81} & \textcolor{blue}{12.72} & \textcolor{blue}{08.32} & \textcolor{blue}{22.30} & \textcolor{blue}{15.50} & \textcolor{blue}{10.79} & \textcolor{blue}{05.18} & \textcolor{blue}{15.89} & \textcolor{blue}{14.67} & \textcolor{blue}{12.46} & \textcolor{blue}{11.77} & \textcolor{blue}{15.45} & \textcolor{blue}{13.55} & \textcolor{blue}{11.53} & \textcolor{blue}{10.61} \\
		RU-RGB* \cite{furnari2019would} & 22.21 & 17.81 & 12.72 & 08.32 & 22.30 & 15.50 & 10.79 & 05.18 & 15.89 & 14.67 & 12.46 & 11.77 & 15.45 & 13.55 & 11.53 & 10.61 \\
		CNN model \cite{abu2018will} & 21.24 & 19.03 & 15.98 & 09.87 & 29.14 & 20.14 & 17.46 & 10.86 & 17.90 & 16.35 & 15.37 & 14.54 & 22.44 & 20.12 & 19.69 & 18.76 \\
		Grammar-based \cite{richard2017weakly} & 24.73 & 22.34 & 19.76 & 12.74 & 29.65 & 19.18 & 15.17 & 13.14 & 16.60 & 14.95 & 13.47 & 13.42 & 21.10 & 18.18 & 17.46 & 16.30 \\
		Uncertainty-based \cite{farha2019uncertainty} & 24.86 & 22.37 & 19.88 & 12.82 & 29.10 & 20.50 & 15.28 & 12.31 & 16.71 & 15.40 & 14.47 & 14.20 & 20.73 & 18.27 & 18.42 & 16.86 \\
		RNN model \cite{abu2018will} & 30.06 & 25.43 & 18.74 & 13.49 & 30.77 & 17.19 & 14.79 & 09.77 & 18.11 & 17.20 & 15.94 & 15.81 & 21.64 & 20.02 & 19.73 & 19.21 \\
		Time-cond. \cite{ke2019time} & 32.51 & 27.61 & 21.26 & \textbf{15.99} & 35.12 & \textbf{27.05} & \textbf{22.05} & \textbf{15.59} & 18.41 & 17.21 & 16.42 & 15.84 & 22.75 & 20.44 & 19.64 & \textbf{19.75} \\
		\hline %\hline
		\textbf{SRL} & \textbf{37.92} & \textbf{28.79} & \textbf{21.30} & 11.05 & \textbf{37.46} & 24.11 & 17.05 & 09.07 & \textbf{25.57} & \textbf{21.04} & \textbf{18.54} & \textbf{16.03} & \textbf{27.31} & \textbf{23.59} & \textbf{20.83} & 17.32 \\
		\hline
	\end{tabular}
	\label{table:thired-person-results}
	\vspace{-1ex}
\end{table*}

\begin{table*}[h]
	\small
	\renewcommand{\arraystretch}{1.05}
	\setlength{\tabcolsep}{2mm}
	\centering
	\caption{Ablation studies on 50 Salads. Given the baseline model, we explore the validity of each component.}
	\begin{tabular}{l|c|c|c|c|c|c|c|c|c|c|c}
		\hline
		\multirow{2}{*}{Setting} & \multirow{2}{*}{Revision} & \multirow{2}{*}{Reattend} & \multirow{2}{*}{Semantic Context} & \multicolumn{4}{c|}{observed 20\%} & \multicolumn{4}{c}{observed 30\%}\\
		\cmidrule{5-12}
		& & & & 10\% & 20\% & 30\% & 50\% & 10\% & 20\% & 30\% & 50\% \\
		\hline %\hline 
		Baseline &   &   &  & 22.96 & 18.26 & 12.96 & 06.14 & 22.69 & 17.12 & 11.72 & 06.01 \\
		+Rev & \checkmark  &   &   & 27.87 & 22.34 & 17.05 & 09.50 & 31.42 & 20.26 & 13.74 & 06.97 \\
		+Rea &   & \checkmark  &   & 24.08 & 21.30 & 15.62 & 07.57 & 26.28 & 18.06 & 12.75 & 07.24 \\
		+SecCon &   &   & \checkmark  & 24.90 & 18.36 & 13.16 & 07.39 & 28.08 & 18.33 & 13.21 & 07.32 \\
		+Rev \& Rea & \checkmark & \checkmark &  & 36.84 & 26.87 & 19.59 & 10.40 & 33.90 & 23.75 &  14.70 & 07.67  \\
		+Rev \& SecCon &  \checkmark &   & \checkmark  & 30.82 & 23.24 & 18.14 & 09.57 & 36.42 & 23.05 & 14.65 & 08.55 \\
		+Rea \& SecCon &   & \checkmark & \checkmark  & 31.49 & 26.66 & 20.80 & 09.57 & 32.26 & 21.06 & 15.94 & 08.66 \\
		\hline %\hline 
		\textbf{SRL} & \checkmark & \checkmark & \checkmark & \textbf{37.92} & \textbf{28.79} & \textbf{21.30} & \textbf{11.05} & \textbf{37.46} & \textbf{24.11} & \textbf{17.05} & \textbf{09.07} \\
		\hline
	\end{tabular}
	\label{table:50salads-ablation-rgb}
	%\vspace{-2ex}
\end{table*}

\subsection{Experiments on Third-person Video}
\label{sec:exp-third}

\subsubsection{Comparison with Other Methods}
\label{sec:exp-third-compare}

In order to verify the generality of SRL, we also conduct experiments on third-person video datasets 50 Salads and Breakfast. We compare SRL with six third-person activity anticipation methods: Nearest-Neighbor, CNN model~\cite{abu2018will}, Grammar-based~\cite{richard2017weakly}, Uncertainty-based~\cite{farha2019uncertainty}, RNN model~\cite{abu2018will} and Time-cond.~\cite{ke2019time}. 
%\textcolor{blue}
We also compare SRL with the state-of-the-art egocentric video activity anticipaiton method RU-RGB~\cite{furnari2019would}. The experimental results are shown in Table~\ref{table:thired-person-results}. 

We can clearly see from Table~\ref{table:thired-person-results} that SRL outperforms most existing methods on Breakfast. On 50 Salads, in addition to individual prediction moments, SRL also achieves better activity anticipation results than existing methods. We can also find that the performance of Time-cond. model is better than SRL at some longer anticipation times. This is most likely because the Time-cond. introduces a time parameter $t$, which denotes the anticipation times. Specifically, the time parameter $t$ is fed to an MLP network to produce a time representation. Then, the time representation and representations of each observed time-step are combined for further processing. This explicit modeling of anticipation time improves the performance of their models for long-term prediction. 

We can find poor anticipation performance for 50\% anticipation on 50 Salads from Table~\ref{table:thired-person-results}. We think this is mainly due to the unique characteristics of videos in the 50 Salads. First, the length of the video in the 50 Salads varies from more than 4 minutes to more than 10 minutes. Second, there are some background frames that do not contain activity information. These two factors pose great challenges to the representation revision and dynamically reattending operations in our model. Hence, given a long observed video clip, the performance of SRL for predicting activities over an exceedingly long period may degrade. 
%\textcolor{blue}

Besides, when we focus on the performance differences between RU-RGB and SRL in Table~\ref{table:epic-results} and Table~\ref{table:thired-person-results}, we can find that SRL obtains good anticipation performance on both EPIC-Kitchens and longer-term anticipation on 50 Salads and Breakfast, while RU-RGB is less successful on 50 Salads and Breakfast. This observation indicates the strong predictive performance of SRL on both egocentric and third-person videos.

%\textcolor{blue}
\subsubsection{Ablation Study on 50 Salads}
\label{sec:exp-third-ablation}

Since the egocentric and third-person activity ancipitation tasks are very different due to the time window they cover (the egocentric anticipation tasks are only focused on the immediate next few seconds rather than the entire rest of the video), we also conduct ablation studies on 50 Salads to see the efficiency of each component of SRL. The results are shown in Table~\ref{table:50salads-ablation-rgb}. 

From the table we can obtain several important conclusions. First, since the third-person datasets do not provide the action and object annotations, we derive the action and object labels from the provided activity annotations. The performance degradation from `SRL' to `+Rev \& Rea'~(or from `+SecCon' to `Baseline') indicates that the prediction of activity-related actions and objects is also necessary for third-person anticipation task. Second, the anticipation performance has different degrees of improvement by adding different single module or any two modules of our SRL to baseline model. These experimental results demonstrate the necessity and the validity of each component of SRL. Therefore, our specific designed future activity anticipation framework for egocentric videos is also effective for third-person videos. Third, by comparing the ablation study results on egocentric and third-person video datasets in Table~\ref{table:epic-ablation-rgb} and Table~\ref{table:50salads-ablation-rgb}, we can find that the most effective component of our SRL, which is `Rev' on 50 Salads and `Rea' (or `SecCon') on EPIC-Kitchens, is different for egocentric and third-person ancipitation tasks. Actually, for third-person anticipation task, it has longer anticipation time window, which may introduce more error accumulation for recursive sequence prediction paradigm. Hence, the `Rev' module of our SRL will be more significant compared with other components, which can be used to improve the representational ability of the predicted intermediate feature and bring greater performance improvements.

%!TEX root = ../dissertation.tex

% Conclusion
\section{Weakness}
\label{sec:weakness}

Our approach performs fairly well in dealing with egocentric video datasets and third-person video datasets, as shown in the experimental results, but there still are several issues to address.
	\begin{itemize}
		\item At each anticipation time, our approach can only give one certain prediction. Since the future is uncertain, like~\cite{yang2019structured}, it would be better for our model to produce multiple predictions and give different confidence values.
		\item When our approach processes long videos with a large number of frames without useful information, the long-term anticipation performance is not as good as the short-term anticipation performance.		
		%\item Our sampling methods for representation revision operation cannot avoid sampling hard negative samples of specific categories, which may limit the training efficacy of the model. 
		\item Our sampling methods for representation revision operation cannot avoid sampling certain types of negative samples. The activity categories of these negative samples are unlikely to co-occur in the same video clip with the activity category of the positive sample, which may provide misleading information and lead to degradation of the training efficacy. 		
		\item In order to better solve the error accumulation problem of the recursive sequence prediction paradigm and make full use of the semantic context contained in the video, we need to choose different values of $\alpha$ and $\beta$ for different datasets and feature modalities, which leads to slight increase of the complexity of our method.
\end{itemize}

%!TEX root = ../dissertation.tex

% Conclusion
\section{Conclusion}
\label{sec:conclusion}

% Future activity anticipation, which aims to predict what possibly happens within a long time horizon, is a challenging problem. In this paper, we analyze three crucial factors we need to consider for building a more perceptive anticipation framework, which are (\uppercase\expandafter{\romannumeral1}) how to accurately rectify predicted intermediate feature representation, (\uppercase\expandafter{\romannumeral2}) how to adaptively attend to observation information according to the predicted representation, (\uppercase\expandafter{\romannumeral3}) how to efficiently utilize semantic context information related to the target activity. Focusing on the three core elements, we propose a novel self regulated learning framework~(SRL) for egocentric video activity anticipation, which considers the three core elements mentioned above in a recursive sequence prediction model. By fully considering and utilizing these three factors, we can finally obtain accurate prediction results. We carry out extensive experiments on egocentric video datasets~(EPIC-Kitchens and EGTEA Gaze+) to verify the effectiveness of our proposed model. Besides, we also perform experiments on third-person video datasets~(50Salads and Breakfast) to prove the universality of our model. In the future, we will extend our network to other tasks, like the pedestrian trajectory prediction.

%As a standard future activity anticipation paradigm, recursive sequence prediction suffers from the accumulation of errors. To address this problem, 

We have proposed an effective Self-Regulated Learning~(SRL) framework to solve the error accumulation problem of recursive sequence prediction pattern for future activity anticipation. SRL aims to regulate the anticipated intermediate representation consecutively to produce more informative representation. Specially, a contrastive loss is utilized to emphasize the novel information in the current anticipation frame in contrast to previously observed content, and a dynamic reweighing mechanism is constructed to exploit the correlation between current frame and previously observed frames, which can attend to informative frames in the observed video clip with a similarity comparison between feature of the current frame and observed frames. Finally, multi-task learning is used to further enhance the learned final video representation, which performs joint feature learning on the target activity labels and the corresponding action and object classes. Experiments on two egocentric video datasets and two third-person video datasets have demonstrated the outstanding performance and effectiveness of the proposed approach. In the future, we will extend our method to other tasks, like the pedestrian trajectory prediction.

\ifCLASSOPTIONcompsoc
\section*{Acknowledgments}
\else
\section*{Acknowledgment}
\fi

The authors would like to thank the associate editor and the reviewers for their time and effort provided to review the manuscript. This work was supported in part by the National Key R\&D Program of China under Grant 2018AAA0102003, in part by National Natural Science Foundation of China: 62022083, 61672497, 61620106009, 61836002 and 61931008, in part by Key Research Program of Frontier Sciences, CAS: QYZDJ-SSW-SYS013, in part by the Beijing Nova Program under Grant Z201100006820023, and in part by the Fundamental Research Funds for the Central Universities. We acknowledge Kingsoft Cloud for the helpful discussion and free GPU cloud computing resource support.

\ifCLASSOPTIONcaptionsoff
\newpage
\fi

%% -----------------------------------------------------------------------------------------------
\bibliographystyle{IEEEtran}
\bibliography{IEEEabrv,reference}

\begin{IEEEbiography}[{\includegraphics[width=1in,height=1.25in,clip,keepaspectratio]{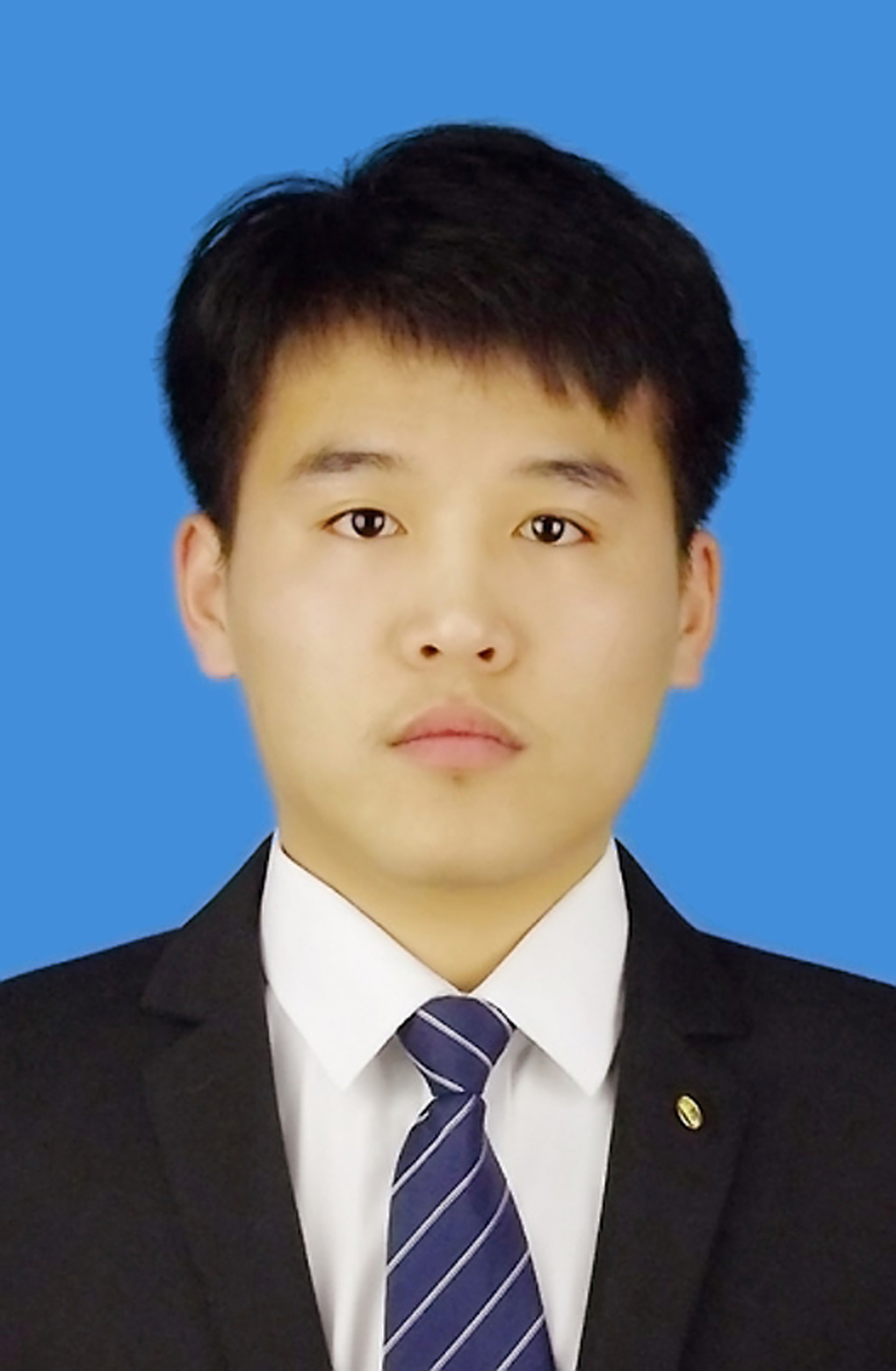}}]{Zhaobo Qi} received the B.S. degree from Harbin Institute of Technology at Weihai in 2016. He is currently pursuing the Ph.D. degree in the School of Computer Science and Technology, University of Chinese Academy of Sciences. His current research interests include video understanding, knowledge engineering and computer vision.
\end{IEEEbiography}

\begin{IEEEbiography}[{\includegraphics[width=1in,height=1.25in,clip,keepaspectratio]{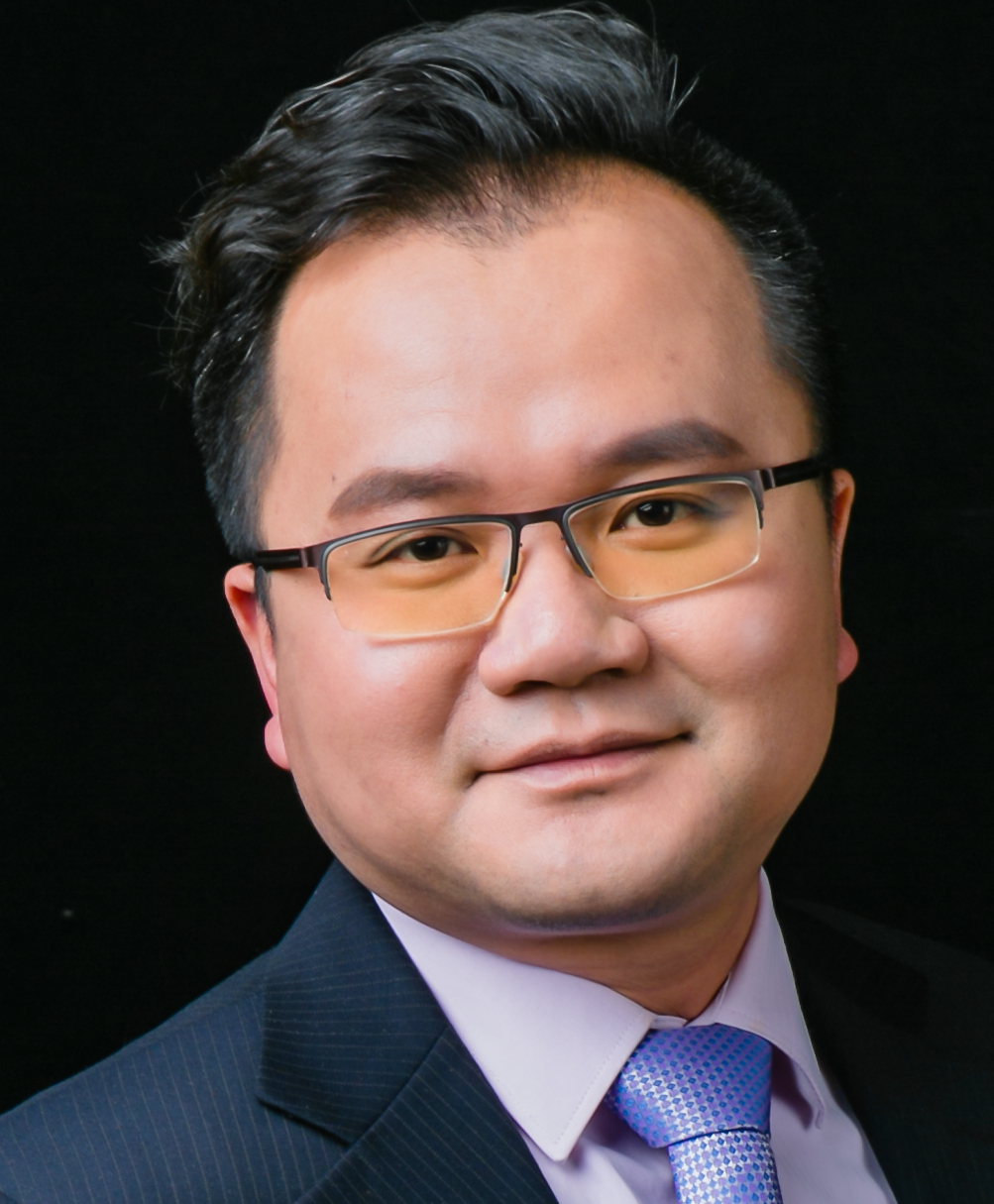}}]{Shuhui Wang} received the B.S. degree in electronics
engineering from Tsinghua University, Beijing, China, in 2006, and the Ph.D. degree from the Institute of Computing Technology, Chinese Academy
of Sciences, Beijing, China, in 2012. He is currently a Full Professor with the Institute of Computing Technology, Chinese Academy of Sciences.
He is also with the Key Laboratory of Intelligent Information Processing, Chinese Academy of Sciences. His research interests include image/video understanding/retrieval, cross-media analysis and visual-textual knowledge extraction.
\end{IEEEbiography}

\begin{IEEEbiography}[{\includegraphics[width=1in,height=1.25in,clip,keepaspectratio]{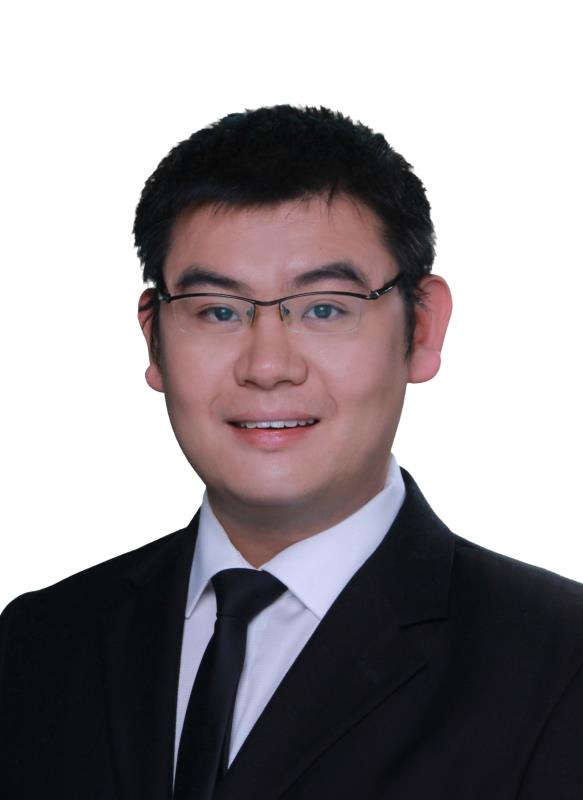}}]{Chi Su} is currently a General Manager of Artificial Intelligence Product Center at Kingsoft Cloud, Beijing. He received the PhD degree in the Institute of Digital Media, EECS, Peking University. His research include computer vision and machine learning, with focus on object detection, object tracking, and human identification and recognition.
\end{IEEEbiography}

\begin{IEEEbiography}[{\includegraphics[width=1in,height=1.25in,clip,keepaspectratio]{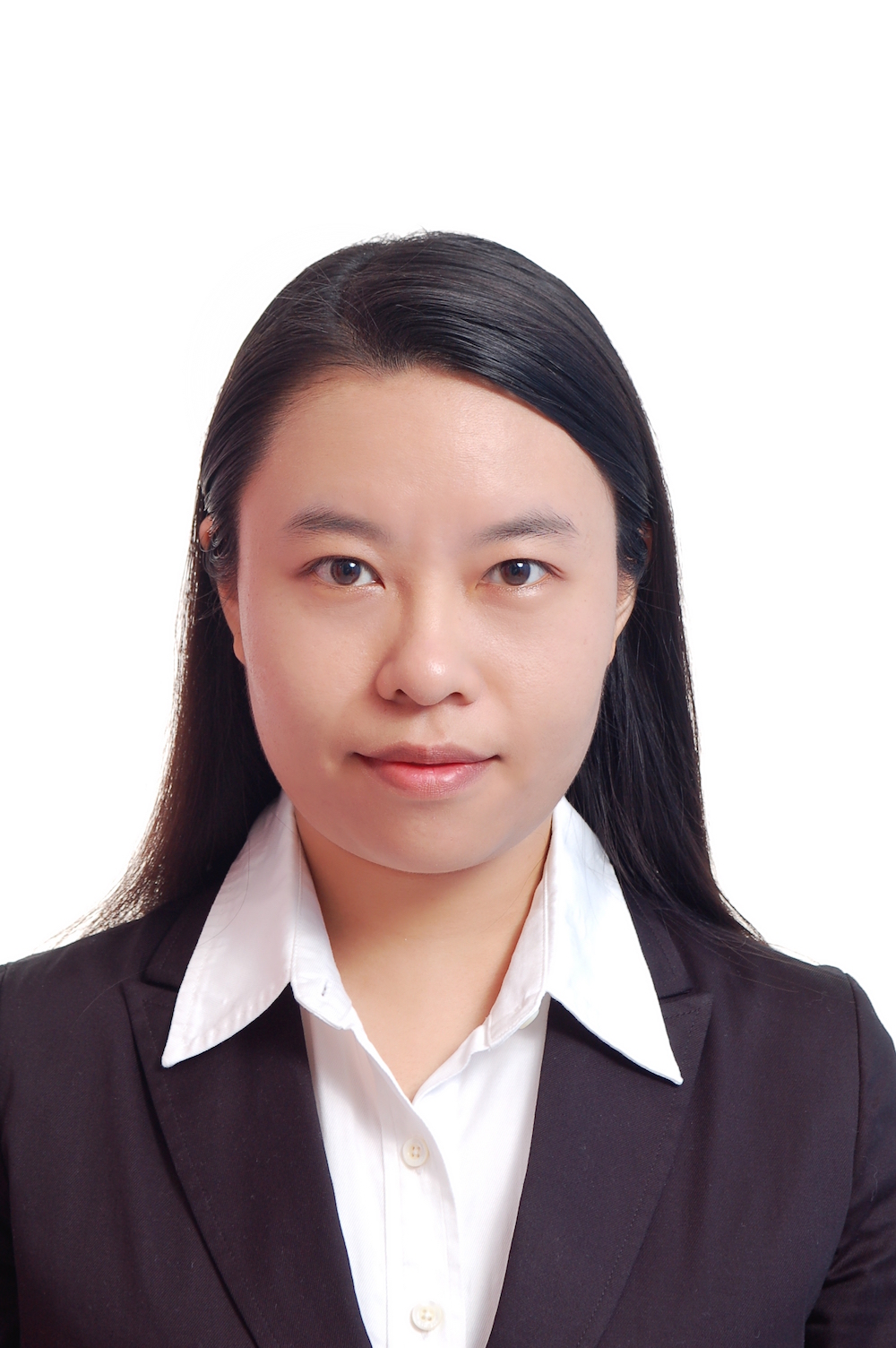}}]{Li Su} received the Ph.D. degree in computer science from the Graduate University of Chinese Academy of Sciences, Beijing, in 2009. She is currently a Full Professor with the School of Computer Science and Technology, University of Chinese Academy of Sciences, Beijing, China. Her research interests include image processing and media computing.
\end{IEEEbiography}

\begin{IEEEbiography}[{\includegraphics[width=1in,height=1.25in,clip,keepaspectratio]{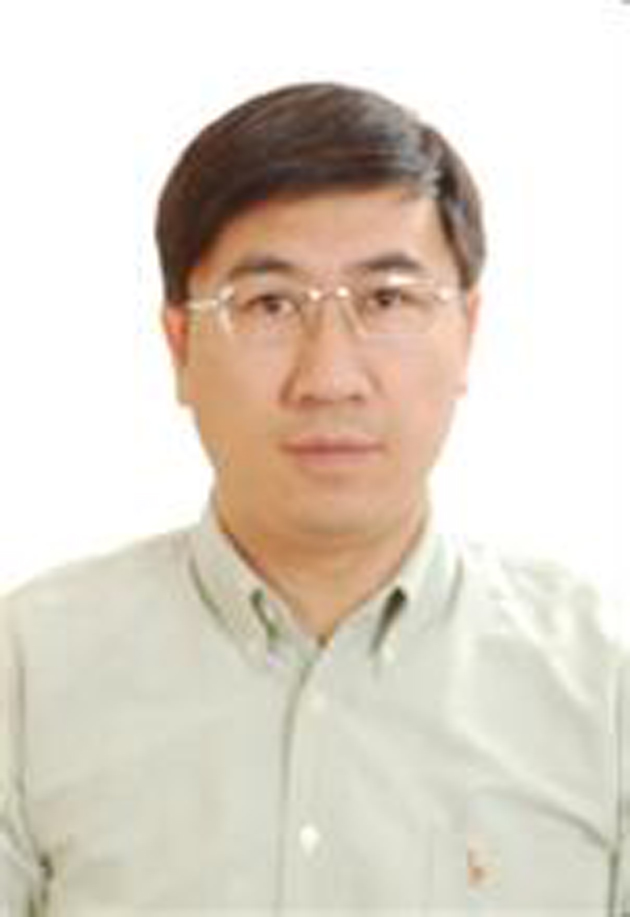}}]{Qingming Huang} received the B.S. degree
in computer science and Ph.D. degree in computer engineering from the Harbin Institute of Technology, Harbin, China, in 1988 and 1994,
respectively.
He is currently a Chair Professor with the School of Computer Science and Technology, University of Chinese Academy of Sciences. He has authored over 400 academic papers in international journals, such as IEEE Transactions on Pattern Analysis and Machine Intelligence, IEEE Transactions on Image Processing, IEEE Transactions on Multimedia, IEEE Transactions on Circuits and Systems for Video Technology, and top level international conferences, including the ACM Multimedia,
ICCV, CVPR, ECCV, VLDB, and IJCAI. He is the Associate Editor of IEEE Transactions on Circuits and Systems for Video Technology and the Associate Editor of Acta Automatica Sinica.
His research interests include multimedia computing, image/video processing, pattern recognition, and computer vision.
\end{IEEEbiography}

\begin{IEEEbiography}[{\includegraphics[width=1in,height=1.25in,clip,keepaspectratio]{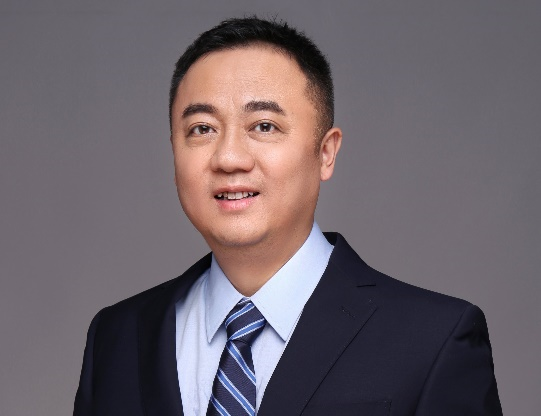}}]{Qi Tian} is currently a Chief Scientist in Artificial Intelligence at Cloud BU, Huawei. From 2018-2020, he was the Chief Scientist in Computer Vision at Huawei Noah's Ark Lab.  He was also a Full Professor in the Department of Computer Science, the University of Texas at San Antonio (UTSA) from 2002 to 2019. During 2008-2009, he took one-year Faculty Leave at Microsoft Research Asia (MSRA). 
Dr. Tian received his Ph.D. in ECE from University of Illinois at Urbana-Champaign (UIUC) and received his B.E. in Electronic Engineering from Tsinghua University and M.S. in ECE from Drexel University, respectively. Dr. Tian's research interests include computer vision, multimedia information retrieval and machine learning and published 590+ refereed journal and conference papers. His Google citation is over 26100+ with H-index 78. He was the co-author of best papers including IEEE ICME 2019, ACM CIKM 2018, ACM ICMR 2015, PCM 2013, MMM 2013, ACM ICIMCS 2012, a Top 10\% Paper Award in MMSP 2011, a Student Contest Paper in ICASSP 2006, and co-author of a Best Paper/Student Paper Candidate in ACM Multimedia 2019, ICME 2015 and PCM 2007.
Dr. Tian research projects are funded by ARO, NSF, DHS, Google, FXPAL, NEC, SALSI, CIAS, Akiira Media Systems, HP, Blippar and UTSA. He received 2017 UTSA President’s Distinguished Award for Research Achievement, 2016 UTSA Innovation Award, 2014 Research Achievement Awards from College of Science, UTSA, 2010 Google Faculty Award, and 2010 ACM Service Award. He is the associate editor of IEEE TMM, IEEE TCSVT, ACM TOMM, MMSJ, and in the Editorial Board of Journal of Multimedia (JMM) and Journal of MVA.  Dr. Tian is the Guest Editor of IEEE TMM, Journal of CVIU, etc. Dr. Tian is a Fellow of IEEE.
\end{IEEEbiography}

\vfill

\end{document}